\pdfoutput=1
\documentclass{elsarticle}
\usepackage{lineno,hyperref}
\usepackage[font=footnotesize,labelfont=bf]{caption}
\usepackage[font=footnotesize,labelfont=bf]{subcaption}
\usepackage{amsmath}
\usepackage{float}
\usepackage{booktabs}
\usepackage{changepage}
\usepackage{xcolor}
\usepackage{adjustbox,lipsum}
\usepackage{multirow}
\usepackage{array}
\newcolumntype{L}[1]{>{\raggedright\let\newline\\\arraybackslash\hspace{0pt}}m{#1}}
\newcolumntype{C}[1]{>{\centering\let\newline\\\arraybackslash\hspace{0pt}}m{#1}}
\newcolumntype{R}[1]{>{\raggedleft\let\newline\\\arraybackslash\hspace{0pt}}m{#1}}
\usepackage[flushleft]{threeparttable} 
\usepackage{graphicx}
\usepackage{rotating}
\definecolor{}{rgb}{0,0,1}
\definecolor{ao}{rgb}{0.0, 0.0,0.0} 

\begin{document}\sloppy
	
	\begin{frontmatter}
		
		\title{SUM: A Benchmark Dataset of \textbf{S}emantic \textbf{U}rban \textbf{M}eshes}
		
		\author[TUDelftaddress]{Weixiao GAO\corref{mycorrespondingauthor}}
		\cortext[mycorrespondingauthor]{Corresponding author}
		\ead{w.gao-1@tudelft.nl}
		
		\author[TUDelftaddress]{Liangliang Nan}
		\ead{liangliang.nan@tudelft.nl}
		
		\author[CMTaddress]{Bas Boom}
		\ead{bboom@cyclomedia.com}
		
		\author[TUDelftaddress]{Hugo Ledoux}
		\ead{h.ledoux@tudelft.nl}
		
		\address[TUDelftaddress]{3D Geoinformation Research Group, Faculty of Architecture and the Built Environment, Delft University of Technology, 2628 BL Delft, The Netherlands}
		\address[CMTaddress]{CycloMedia Technology, Zaltbommel, The Netherlands}
		
		\begin{abstract}
		Recent developments in data acquisition technology allow us to collect 3D texture meshes quickly.
		Those can help us understand and analyse the urban environment, and as a consequence are useful for several applications like spatial analysis and urban planning.
		Semantic segmentation of texture meshes through deep learning methods can enhance this understanding, but it requires a lot of labelled data.
		The contributions of this work are three-fold: (1) a new benchmark dataset of semantic urban meshes, (2) a novel semi-automatic annotation framework, and (3) an annotation tool for 3D meshes.
		In particular, our dataset covers about  4 $km^2$ in Helsinki (Finland), with six classes, and we estimate that we save about 600 hours of labelling work using our annotation framework, which includes initial segmentation and interactive refinement.
		We also compare the performance of several state-of-the-art 3D semantic segmentation methods on the new benchmark dataset.  
		Other researchers can use our results to train their networks: the dataset is publicly available, and the annotation tool is released as open-source.
		\end{abstract}
	
	\end{frontmatter}
	
	\section{Introduction}%
\label{sec:intro}

Understanding the urban environment from 3D data (e.g.\ point clouds and 3D meshes) is a long-standing goal in photogrammetry and computer vision~\citep{matrone2020heritage, hackel2017semantic3d}. 
The fast recent developments in data acquisition technologies and processing pipelines have allowed us to collect a great number of datasets on our 3D urban environments. 
Prominent examples are Google Earth~\citep{Google3d}, texture meshes covering entire cities (e.g.\ Helsinki~\citep{Helsinki3d}), or point clouds covering entire countries (e.g., the Netherlands AHN~\citep{ahn2019}). 
These datasets have attracted interest because of their potential in several applications, for instance, urban planning~\citep{Chen2011,czynska2014application}, positioning and navigation~\citep{cappelle2012virtual,Peyraud2013,Hsu2015}, spatial analysis~\citep{Yaagoubi2015}, environmental analysis~\citep{Deng2016}, and urban fluid simulation~\citep{GarciaSanchez14}. 

\begin{figure}[H]
	\includegraphics[width=\linewidth]{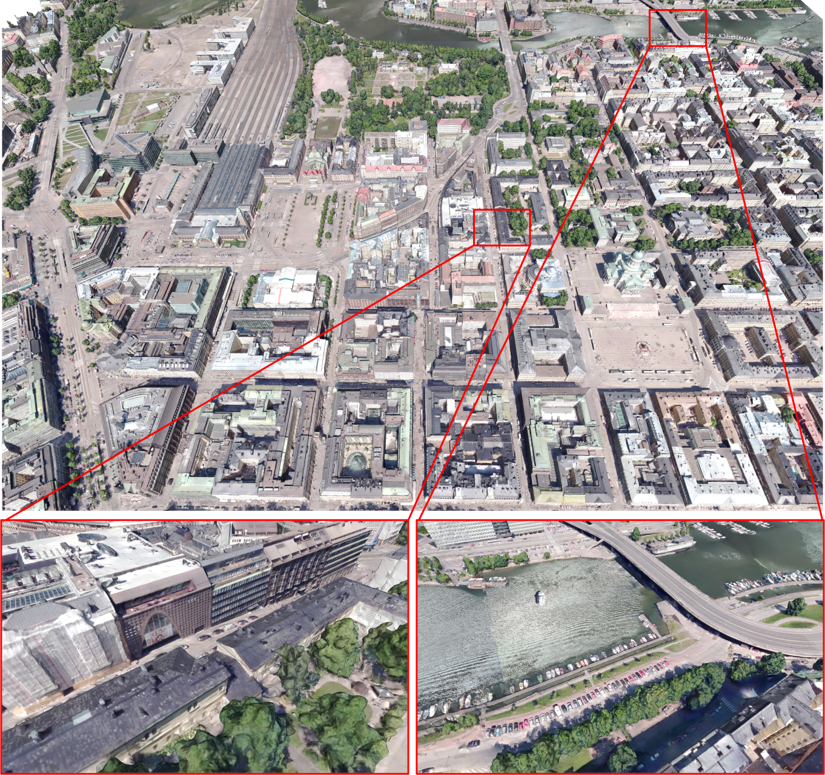} 
	\caption{Part of the semantic urban mesh benchmark dataset shown as a texture mesh.}
	\label{fig:texside}
\end{figure}

\begin{figure}[H]
	\includegraphics[width=\linewidth]{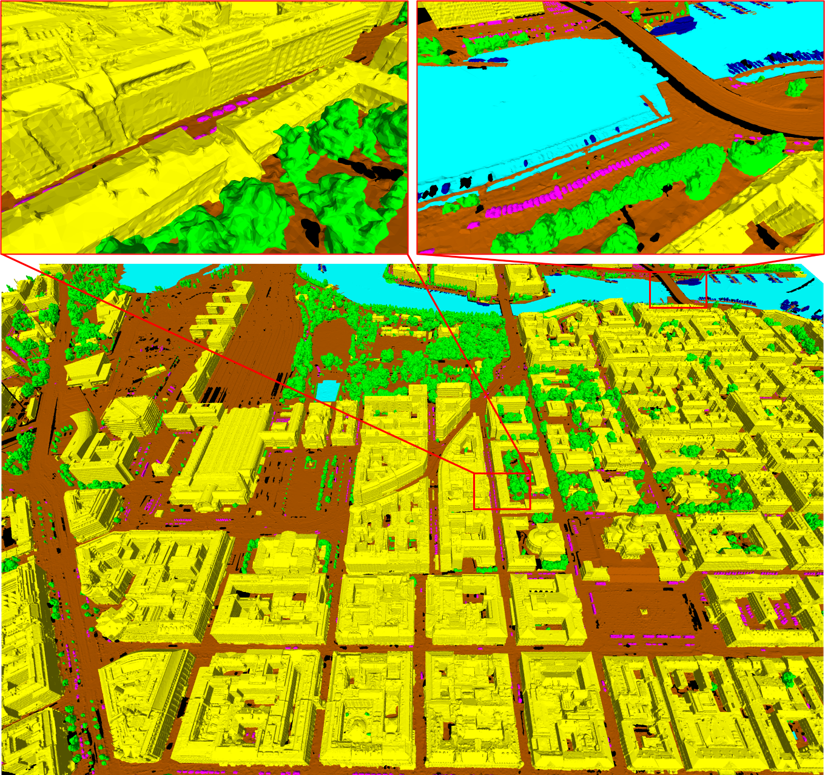} 
	\centering
	\includegraphics[width=0.9\textwidth]{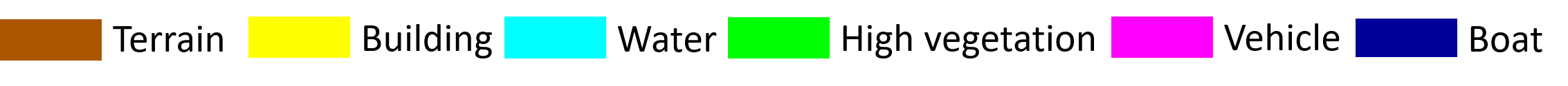}
	\caption{Part of the semantic urban mesh benchmark dataset, showing the semantic classes (unclassified regions are in black). }
	\label{fig:semside}
\end{figure}

To effectively understand the urban phenomena behind the data, a large amount of ground truth is typically required, especially when applying supervised learning-based techniques, such as a deep Convolutional Neural Network (CNN). 
The recent development of machine learning (especially deep learning) techniques has demonstrated promising performance in semantic segmentation of 3D point clouds~\citep{qi2017pointnet,landrieu2018large, thomas2019kpconv}. 
Compared to point clouds, a surface representation (in the form of a 3D mesh, often with textures, see Figure~\ref{fig:texside} and \ref{fig:semside} for an example) of the urban scene has multiple advantages: easy to acquire, compact storage, accurate, and with well-defined topological structures.

This means that 3D meshes have the potential to serve as input for scene understanding. 
As a consequence, there is an urgent demand for large-scale urban mesh datasets that can be used as ground truth for both training and evaluating the 3D semantic segmentation workflows.

In this paper, we aim to establish a benchmark dataset of large-scale urban meshes reconstructed from aerial oblique images. 
To achieve this goal, we propose a semi-automatic mesh annotation framework that includes two components: (1) an automatic process to generate intermediate labels from the raw 3D mesh; (2) manual semantic refinement of those labels.
For the intermediate label generation step, we have developed a semantic mesh segmentation method that classifies each triangle into a pre-defined object class. 
This semantic initialization allows us to achieve an overall accuracy of 93.0\% in the classification of the triangle faces in our dataset, saving significant efforts for manually labelling.
Then, in the semantic refinement step, a mesh annotation tool (which we have developed) is used to refine the semantic labels of the pre-labelled data (at the triangle and segment levels).

We have used our proposed framework to generate a semantic-rich urban mesh dataset consisting of 19 million triangles and covering about 4 $km^2$ with six object classes commonly found in an urban environment: terrain, high-vegetation, building, water, vehicle, and boat (Figure~\ref{fig:semside} shows an example from our dataset).
With our semi-automatic annotation framework, generating the ground truth took only about 400 hours; we estimate that manually labelling the triangles would have taken more than 1000 hours. 
The contributions of our work are: 
\begin{itemize}
	\item a semantic-rich urban mesh dataset of six classes of common urban objects with texture information;
	\item a semi-automatic mesh annotation framework consisting of two parts: a pipeline for semantic mesh segmentation and 
	an annotation tool for semantic refinement;
	\item a comprehensive evaluation and comparison of the state-of-the-art semantic segmentation methods on the new dataset.
\end{itemize}
The benchmark dataset is freely available, and the semantic mesh segmentation methods and the annotation software for 3D meshes are released as open-source\footnote{\url{https://3d.bk.tudelft.nl/projects/meshannotation/}}.

	\section{Related Work}
\label{sec:relaw}

\begin{sidewaystable}
\noindent\adjustbox{max width=\textwidth}
{
\begin{threeparttable}
	\begin{tabular}[t]{@{}lcccC{0.15\linewidth}ccC{0.16\linewidth}C{0.15\linewidth}C{0.2\linewidth}C{0.2\linewidth}c@{}} 
	\toprule
	Name && Platforms & Year & Data Type & Area\tnote{a} / Length & Classes & Points / Triangles & RGB & Automatic Pre-labelling & Annotation & Time Cost (hours)\\
	\midrule
	\textbf{Oakland 3D}~\citep{munoz2009contextual}        && MLS & 2009 & Point Cloud & 1.5 $km$  & 5  & 1.6 $ M $  & No  & No                       & 3D Manually            & Not reported\\
	\textbf{Paris-rue-Madame}~\citep{serna2014paris}       && MLS & 2014 & Point Cloud & 0.16 $km$ & 17 & 20 $ M $   & No  & 2D semantic segmentation & 3D Semi-manually       & Not reported\\	
	\textbf{iQmulus}~\citep{vallet2015terramobilita}       && MLS & 2015 & Point Cloud & 10 $km$   & 8  & 300 $ M $  & No  & No                       & 2D Semi-manually       & Not reported\\
	\textbf{Semantic3D}~\citep{hackel2017semantic3d}       && TLS & 2017 & Point Cloud & -        & 8  & 4000 $ M $ & Yes & No                       & 2D \& 3D Semi-manually  & Not reported\\
	\textbf{Paris-Lille-3D}~\citep{roynard2018paris}       && MLS & 2018 & Point Cloud & 1.94 $km$ & 9  & 143 $ M $  & No  & No                       & 3D Manually            & Not reported\\
	\textbf{SemanticKITTI}~\citep{behley2019semantickitti} && MLS & 2019 & Point Cloud & 39.2 $km$ & 25 & 4549 $ M $ & No  & No                       & 3D Manually            & 1700        \\
	\textbf{Toronto-3D}~\citep{tan2020toronto}             && MLS & 2020 & Point Cloud & 1.0 $km$  & 8  & 78.3 $ M $ & Yes & No                       & 3D Manually            & Not reported\\
	\textbf{ISPRS}~\citep{niemeyer2014contextual}          && ALS & 2012 & Point Cloud & 0.1 $km^2$  & 9  & 1.2 $ M $   & No & No                       & 3D Manually & Not reported\\
	\textbf{AHN3}~\citep{ahn2019}                         && ALS & 2019 & Point Cloud & 41,543 $km^2$ & 4 & 415.43 $ B $\tnote{b} & No & 3D semantic segmentation & 3D Manually & Not reported\\
	\textbf{DublinCity}~\citep{zolanvari2019dublincity}    && ALS & 2019 & Point Cloud & 2.0 $km^2$     & 13 & 260 $ M $             & No & No                       & 3D Manually & 2500        \\
	\textbf{DALES}~\citep{varney2020dales}                 && ALS & 2020 & Point Cloud & 10.0 $km^2$    & 8  & 505.3 $ M $           & No & 3D semantic segmentation & 3D Manually & Not reported\\
	\textbf{LASDU}~\citep{ye2020lasdu}                     && ALS & 2020 & Point Cloud & 1.02 $km^2$    & 5  & 3.12 $ M $            & No & No                       & 3D Manually & Not reported\\
	\textbf{ETHZ RueMonge}~\citep{brostow2009semantic,riemenschneider2014learning} && Auto-mobile camera & 2014 & Mesh & 0.7 $km$ & 9 & 1.8 $ M $ (lowres)\tnote{c} & Yes (per vertex)\tnote{d} & 2D over-segmentation & 2D Semi-manually & 230 (701 frames)\tnote{e} \\
	\textbf{Campus3D}~\citep{li2020campus3d}       && UAV camera & 2020 & Point Cloud & 1.58 $km^2$ & 14 & 937.1 $ M $  & Yes           & No & 2D \& 3D Manually       & Not reported\\
	\textbf{SensatUrban}~\citep{hu2021towards}     && UAV camera & 2020 & Point Cloud & 6 $km^2$    & 13 & 2847.1 $ M $ & Yes           & No & 3D Manually            & 600 \\	
	\textbf{Swiss3DCities}~\citep{can2020semantic} && UAV camera & 2020 & Point Cloud & 2.7 $km^2$  & 5  & 226 $ M $    & Yes           & No & 3D Manually (on mesh)  & 144 (1 $ M $ Triangles)\tnote{f} \\	
	\textbf{Hessigheim 3D}~\citep{laupheimer2020association,kolle2021h3d} && UAV Lidar \& camera & 2021 & Point Cloud \& Mesh & 0.19 $km^2$ & 11 & 125.7 $ M $ / 36.76 $ M $\tnote{g}  & Yes (texture)\tnote{h} & No & 3D Manually\tnote{i} & Not reported\\
	\textbf{SUM-Helsinki (Ours)}                   && Airplane camera & 2021 & Mesh        & 4 $km^2$    & 6  & 19 $ M $     & Yes (texture)\tnote{h} & 3D over-segmentation \& 3D semantic segmentation & 3D Semi-manually  & 400 \\ 
	\bottomrule
	\end{tabular}
  \begin{tablenotes}
  	\item[a] The area was measured in a 2D map.
  	\item[b] The number of total points (i.e., 415.43 billion) is estimated.
	\item[c] The low-resolution meshes contain 1.8 million triangle faces, according to the publications.
	\item[d] An RGB colour was assigned to each triangle vertex.
	\item[e] The frames were from video sequences.
	\item[f] About one million triangles (16 tiles) from simplified mesh were labelled, which took around 6 to 12 hours per tile.
	\item[g] The number of LiDAR points is 125.7 million and the number of triangle faces is 36.76 million.
	\item[h] The colour of each triangle face corresponds to a patch of the texture image.
	\item[i] The LiDAR point clouds were manually annotated and the labels were transferred to the mesh.
  \end{tablenotes}
\end{threeparttable}
}
\caption{Comparison of existing 3D urban benchmark datasets.} 
\label{tab:overview}
\end{sidewaystable}

Urban datasets can be captured with different sensors and be reconstructed with different methods, and the resulting datasets will have different properties.
Most benchmark urban datasets focus on point clouds, whereas our semantic urban benchmark dataset is based on textured triangular meshes.

The input of the semantic labelling process can be raw or pre-labelled urban datasets such as the automatically generated results from over-segmentation or semantic segmentation (see Section~\ref{sec:mesh_annota}).
Regardless of the input data, it still needs to be manually checked and annotated with a labelling tool, which involves selecting a correct semantic label from a predefined list for each triangle (or point, depending on the dataset) by users.
In addition, some interactive approaches can make the labelling process semi-manual.                                                                                                                                        
However, unlike our proposed approach, the labelling work of most of the 3D benchmark data does not take full advantage of over-segmentation and semantic segmentation on 3D data, and interactive annotation in the 3D space.      

We present in this section an overview of the publicly available semantic 3D urban benchmark datasets categorised by sensors and reconstruction types (see Table \ref{tab:overview}).
More specifically, we elaborate on the quality, scale, and labelling strategy of the existing urban datasets regarding semantic segmentation.

\subsection{Photogrammetric Products}

\subsubsection{Dense Point Clouds} The \emph{Campus3D}~\citep{li2020campus3d} is to our knowledge the first aerial point cloud benchmark.
The coarse labelling is conducted in 2D projected images with three views, and the grained labels are refined in 3D with user-defined rotation angles. 
The dataset covers only the campus of the National University of Singapore and is thus not representative of a typical urban scene.

\emph{SensatUrban}~\citep{hu2021towards} is another example of the photogrammetric point clouds covering various urban landscapes in two cities of the UK\@.
The semantic points are manually annotated via the off-the-shelf software tool CloudCompare~\citep{Girardeau-Montaut2016}, and the overall annotation is reported to have taken around 600 hours.
The dataset also contains several areas without points, especially for water surfaces and regions with dense objects. 
The leading causes are the Lambertion surface assumption during the image matching and the inadequate image overlapping rate during the flight.   

Similarly, the \emph{Swiss3DCities}~\citep{can2020semantic} was recently released that covers three cities in Zurich but twice smaller than the SensatUrban.
The annotation work was conducted on a simplified mesh in the software Blender~\citep{blender}, and then the semantics were transferred to the mesh vertices, which are regarded as point clouds, via the nearest neighbour search.    
The mesh simplification may result in the loss of small-scale objects such as building dormers and chimneys, and the automatic transfer of the labels could have introduced errors in the ground truth. 

\subsubsection{Triangle Meshes} To the best of our knowledge, the \emph{ETHZ RueMonge 2014}~\citep{riemenschneider2014learning} is the first urban-related benchmark dataset available as surface meshes.
The label for each triangle is obtained from projecting selected images that are manually labelled from over-segmented image sequences~\citep{brostow2009semantic}. 
In fact, due to the error of multi-view optimisation and the ambiguous object boundary within triangle faces, the datasets contain many misclassified labels, making them unsuitable for training and evaluating supervised-learning algorithms.

\emph{Hessigheim 3D}~\citep{laupheimer2020association,kolle2021h3d} is a small-scale semantic urban dataset consisting of highly dense LiDAR point clouds and high resolution texture meshes.
Particularly, the mesh is generated from both LiDAR point cloud and oblique aerial images in a hybrid way.
The labels of point clouds are manually annotated in CloudCompare~\citep{Girardeau-Montaut2016}, and the labels of the mesh are transferred from the point clouds by computing the majority votes per triangle.
However, if the mesh triangle has no corresponding points, some faces may remain unlabelled which resulted in about 40\% unlabelled area.
In addition, this dataset contains non-manifold vertices, which makes it difficult to use directly. 

\subsection{LiDAR Point Clouds}
Unlike photogrammetric point clouds, LiDAR point clouds usually do not contain colour information.
To annotate them properly, additional information is often required, e.g.\ images or 2D maps.
LiDAR point cloud benchmark datasets are more common than photogrammetric ones.

\subsubsection{Street-view Datasets}
The \emph{Oakland 3D}~\citep{munoz2009contextual} is one of the earliest mobile laser scanning (MLS) point cloud datasets, which was designed for the classification of outdoor scenes. 
It has five hand-labelled classes with 44 sub-classes, but without colour information and semantic categories like roof, canopy, or interior building block, which are typical for all street-view captured datasets.

Compared to \emph{Oakland 3D}, \emph{Paris-rue-Madame}~\citep{serna2014paris} is a relatively smaller dataset which used the 2D semantic segmentation results for 3D annotation.
Specifically, the point clouds were projected onto images to extract the objects hierarchically with several unsupervised segmentation and classification algorithms.

Although the 2D pre-labelled generation is fully automatic, different semantic categories require different segmentation algorithms resulting in difficulties in the classification of multiple classes. 

The \emph{iQmulus dataset}~\citep{vallet2015terramobilita} is a 10 $ km $ street dataset annotated based on projected images in the 2D space. 
Specifically, the user first needs to extract objects by editing the image with a polyline tool and then assigns labels to the extracted object regions.  
Some automatic functions are made for polyline editing in this framework, but the entire annotation pipeline is still complicated.

Unlike other street view datasets, \emph{Semantic3D}~\citep{hackel2017semantic3d} is a dataset consisting of terrestrial laser scanning (TLS) point clouds (the scanner is not moving and scans are made from only a few viewpoints).
It has eight classes and colours were obtained by projecting the points onto the original images.
There are two annotation methods: (1) annotating in 3D with an iterative model-fitting approach on manually selected points; (2) annotating in a 2D view by separate background from a drawn polygon in CloudCompare~\citep{Girardeau-Montaut2016}.
Although it covers many urban scenes and includes RGB information, the acquired objects are incomplete because of the limited viewpoints and occlusions. 

The other three typical MLS point cloud datasets that were manually labelled are \emph{Paris-Lille-3D}~\citep{roynard2018paris}, \emph{SemanticKITTI}~\citep{behley2019semantickitti}, and \emph{Toronto-3D}~\citep{tan2020toronto}.

\subsubsection{Aerial-view Datasets}
As for ALS benchmark point clouds, representative datasets are \emph{ISPRS}~\citep{niemeyer2014contextual}, \emph{DublinCity}~\citep{zolanvari2019dublincity}, and \emph{LASDU}~\citep{ye2020lasdu} covering various scales of city landscapes and were annotated manually with off-the-shelf software. 
Instead of fully manual annotation, the \emph{Dayton Annotated LiDAR Earth Scan (DALES)}~\citep{varney2020dales} used digital elevation models (DEM) to distinguish ground points with a certain threshold, the estimated normal to label the building points roughly, and satellite images to provide contextual information as references for annotators to check and label the rest of data.
Similarly, the AHN3 dataset~\citep{ahn2019} was semi-manually labelled by different companies with off-the-shelf software.
Besides, since the ALS measurement is conducted in the top view direction, unlike oblique aerial cameras, the obtained point clouds often miss facade information to a certain degree.

	\section{The Semantic Urban Mesh Dataset}\label{sec:framework}
\subsection{Dataset Specification}

\begin{figure}[!tb]
	\includegraphics[height=0.48\textwidth]{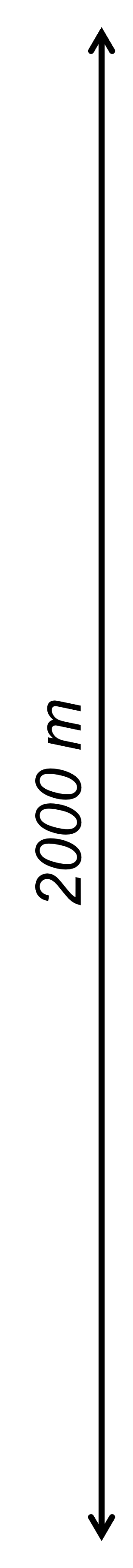}
	\begin{subfigure}[t]{0.48\textwidth}
		\includegraphics[width=\linewidth]{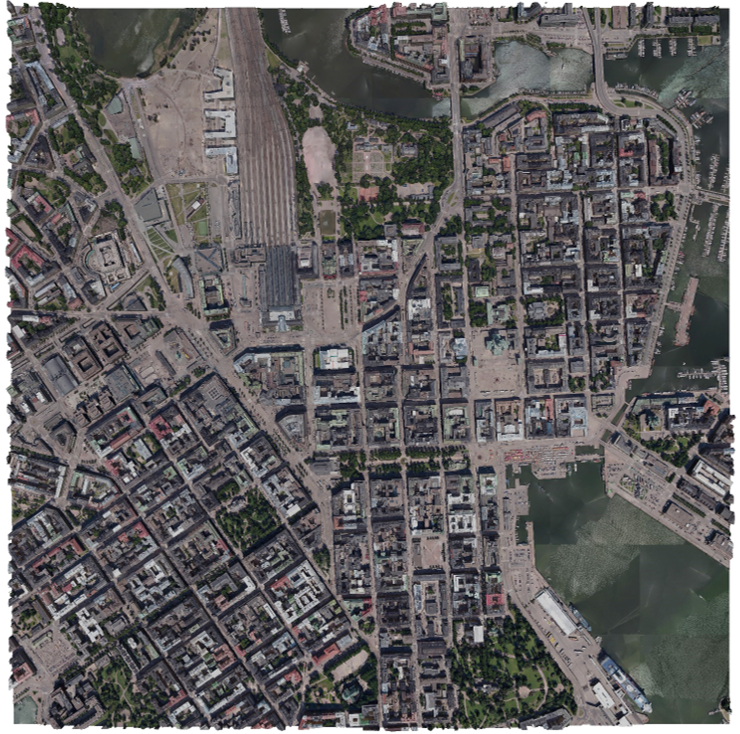}
		\includegraphics[width=\linewidth]{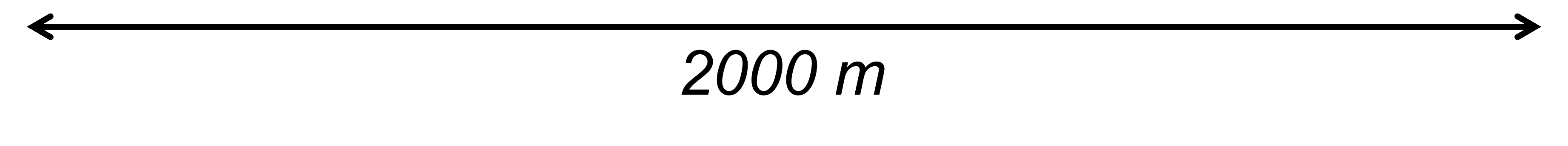}
		\label{fig:textop}
	\end{subfigure}
	\hspace*{\fill}
	\begin{subfigure}[t]{0.48\textwidth}		
		\includegraphics[width=\linewidth]{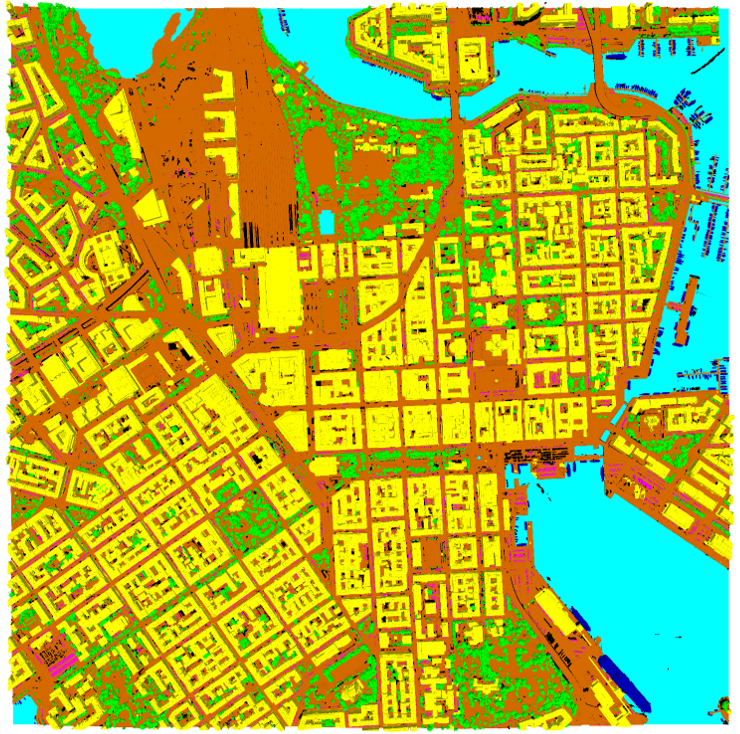}
		\vspace*{-0.78cm}
		\begin{center}
		\includegraphics[width=0.8\linewidth]{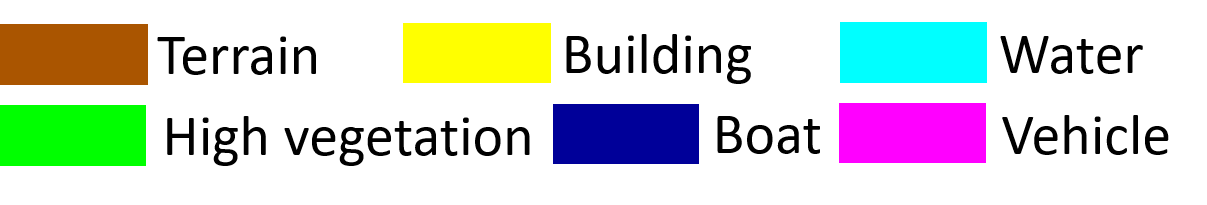}
		\end{center}
		\label{fig:semtop}
	\end{subfigure}
	\vspace*{-0.7cm}
	\caption{Overview of the semantic urban mesh benchmark.
	Left: the texture meshes covering about 4 $km^2$ map area. Right: the ground truth meshes.
	More views of the same scene (with different visualization styles) are shown in Figures \ref{fig:texside} and \ref{fig:semside}.}
	\label{fig:overview}
\end{figure}

We have used Helsinki's 3D texture meshes as input and annotated them as a benchmark dataset of semantic urban meshes. 
The Helsinki's raw dataset covers about 12 $ km^2 $, and it was generated in 2017 from oblique aerial images that have about a 7.5 $cm$  ground sampling distance (GSD) using an off-the-shelf commercial software namely ContextCapture~\citep{contextcap}.
The source images have three colour channels (i.e., red, green, and blue) and are collected from an airplane with five cameras that have $80\%$ length coverage and $60\%$ side coverage.
To recover the 3D water bodies that do not fulfil the Lambertian hypothesis, 2D vector maps and ortho-photos are used when performing the surface reconstruction.
Furthermore, processing like aerial triangulation, dense image matching, and mesh surface reconstruction were all performed with ContextCapture.
It should be noticed that the entire region of Helsinki is split into tiles, and each of them covers about 250 $ m^2 $~\citep{kalasatamaReport}.
As shown in Figure \ref{fig:overview},  we have selected the central region of Helsinki as the study area, which includes 64 tiles and covers about 4 $km^2$ map area (8 $km^2$ surface area) in total.   

\subsection{Object Classes}
\begin{figure}[!tb]
	\centering
	\begin{subfigure}[t]{0.48\textwidth}
		\includegraphics[width=\linewidth]{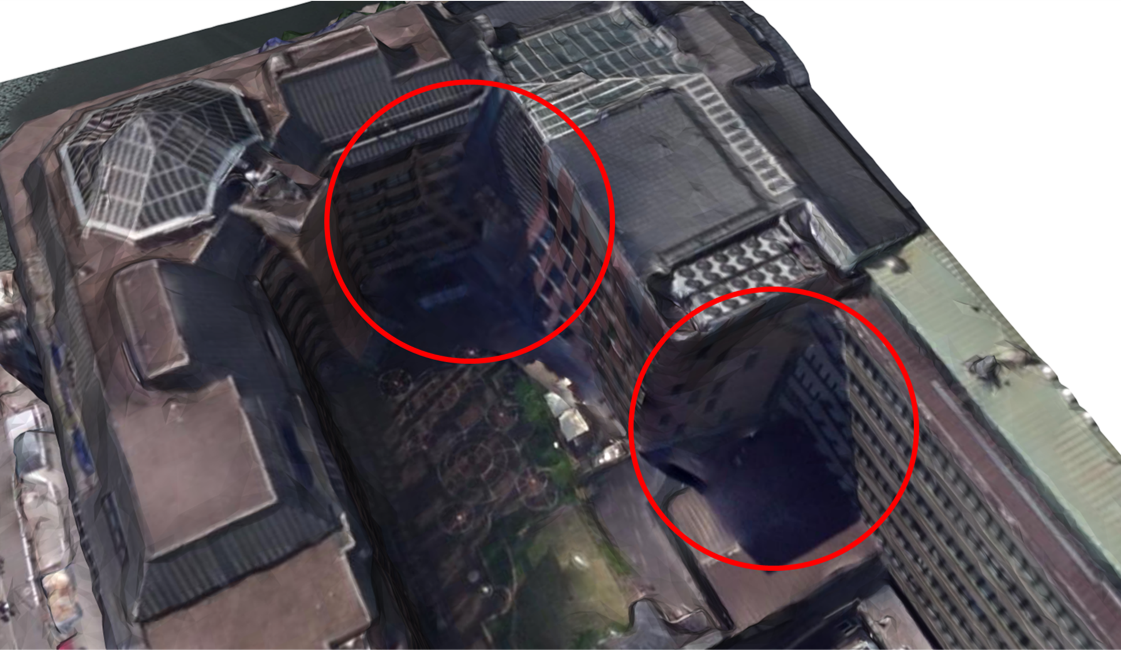}
		\caption{}
	\end{subfigure}
	\hspace*{\fill}
	\begin{subfigure}[t]{0.48\textwidth}
		\includegraphics[width=\linewidth]{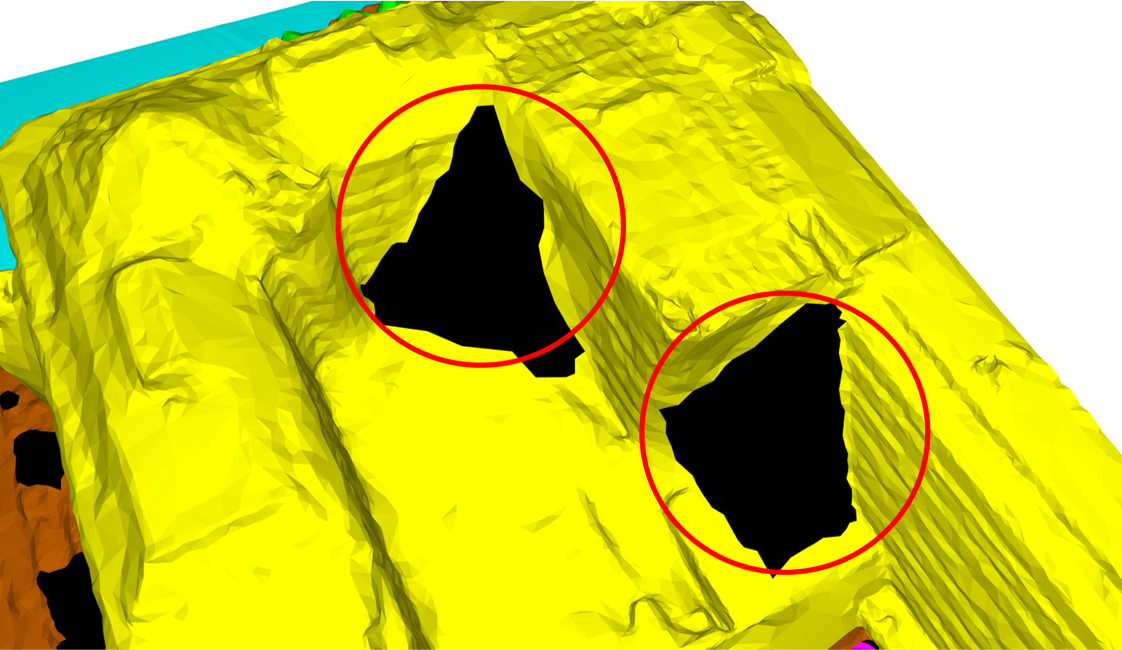}
		\caption{}
	\end{subfigure}
	\begin{subfigure}[t]{0.48\textwidth}
		\includegraphics[width=\linewidth]{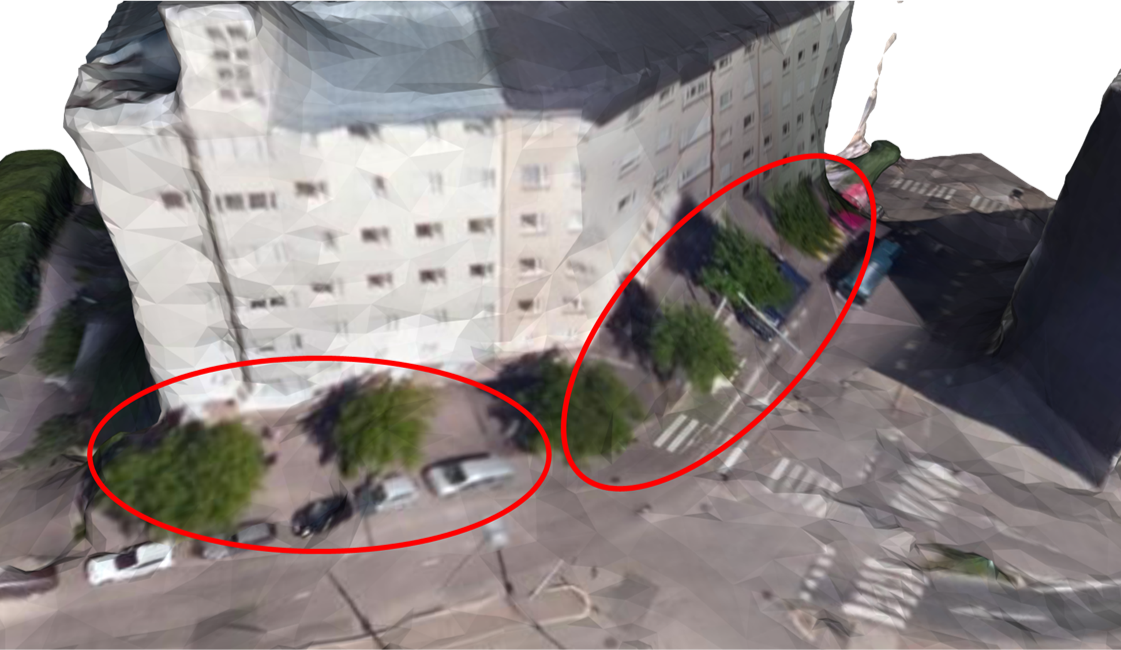}
		\caption{}
	\end{subfigure}
	\hspace*{\fill}
	\begin{subfigure}[t]{0.48\textwidth}
		\includegraphics[width=\linewidth]{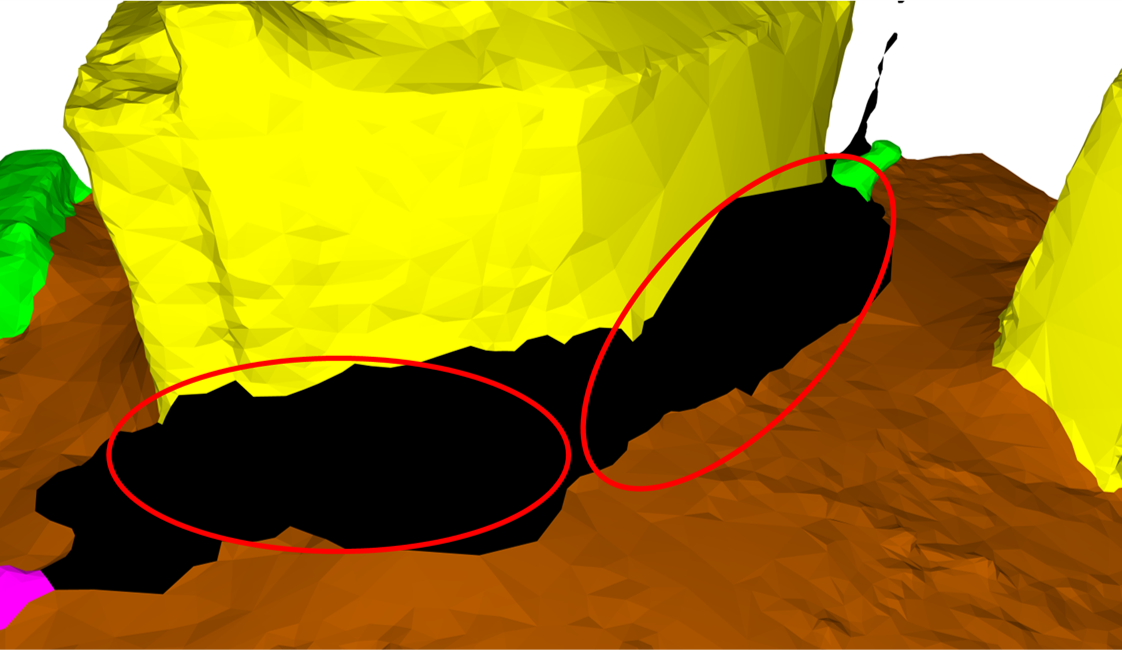}
		\caption{}
	\end{subfigure}
	\caption{Ambiguous regions are labelled as unclassified (in black). 
		(a) Shadow region with texture.
		(b) Shadow region with semantic colour.
		(c) Distorted region with texture.
		(d) Distorted region with semantic colour.} 
	\label{fig:ambigious}
\end{figure}

We define the semantic categories for urban meshes by the most common objects in the urban environment with unambiguous geometry and texture appearance.
Moreover, each triangle face is assigned to a label of one of the six semantic classes. 
Ambiguous regions (which account for about 2.6\% of the total mesh surface area), such as shadowed regions or distorted surfaces, are labelled as unclassified (see Figure \ref{fig:ambigious}).
The object classes we consider in the benchmark dataset are: 
\begin{itemize}
	\item \textbf{terrain}: roads, bridges, grass fields, and impervious surfaces;
	\item \textbf{building}: houses,high-rises, monuments, and security booths;
	\item \textbf{high vegetation}: trees, shrubs, and bushes;
	\item \textbf{water}: rivers, sea, and pools;
	\item \textbf{vehicle}: cars, buses, and lorries;  
	\item \textbf{boat}: boats, ships, freighters, and sailboats;
	\item \textbf{unclassified}: incomplete objects like buses and trains, distorted surfaces like tables, tents and facades, construction sites, underground walls.
\end{itemize}

\subsection{Semi-automatic Mesh Annotation}  \label{sec:mesh_annota}
\begin{figure}[!tb]
	\centering
	\includegraphics[width=\textwidth]{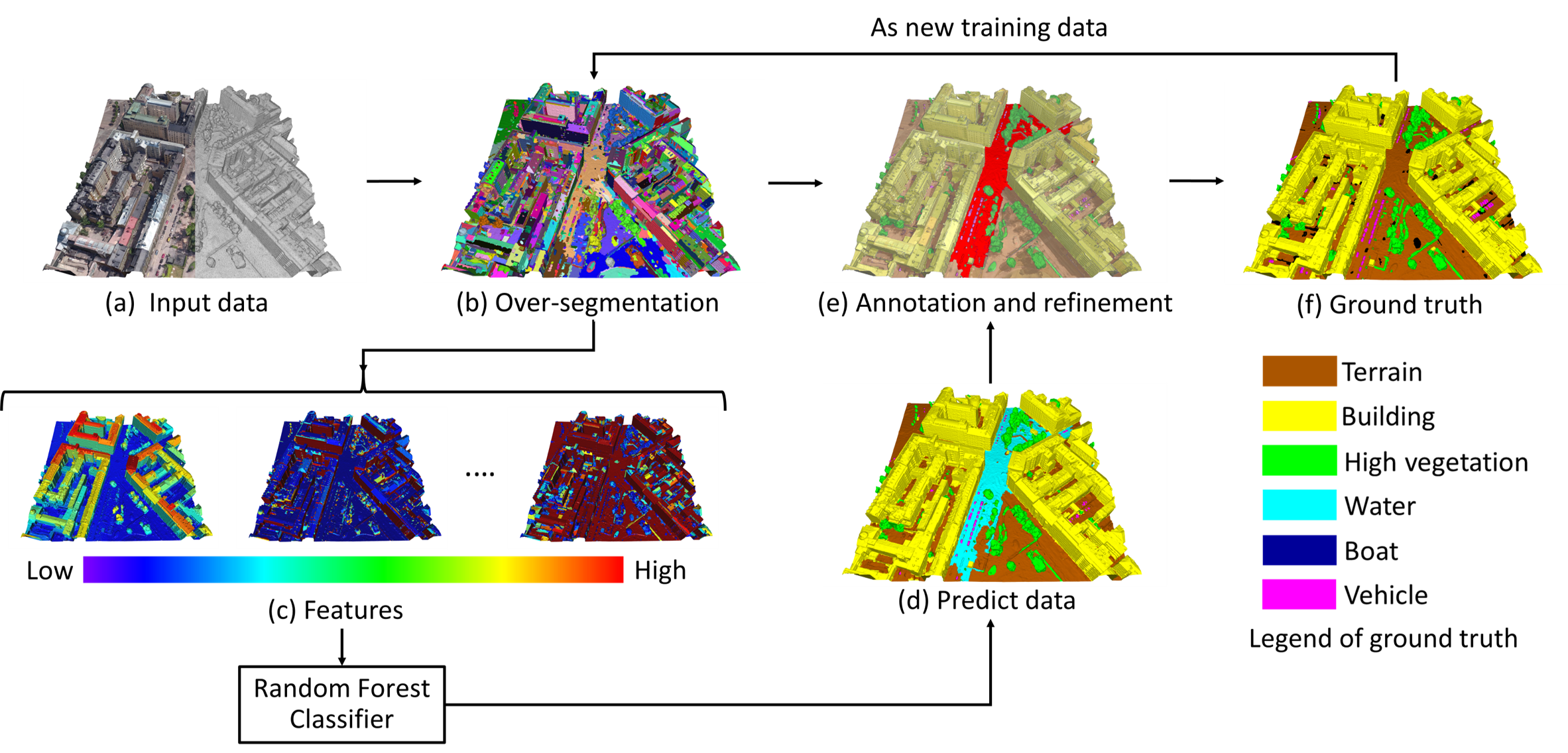}
	\caption{The pipeline of the labelling workflow.}
	\label{fig:pipeline}
\end{figure}

Rather than manually labelling each triangle face of the raw meshes, we design a semi-automatic mesh labelling framework to accelerate the labelling process. Figure~\ref{fig:pipeline} shows the overall pipeline of our labelling workflow.

Given the fact that urban environments consist of a large number of planar regions in the data, we opt to label the data at the segment level instead of individual triangle faces. 
Specifically, we over-segment the input meshes into a set of planar segments. 
These segments can enrich local contextual information for feature extraction and serve as the basic annotation unit to improve annotation efficiency.

Instead of randomly choosing a mesh tile as input for annotation and refinement, which is insufficient for manual annotation progress, we favour picking a mesh tile that is more difficult to classify.
Similar to active learning, we first compute the feature diversity (see Equation \ref{eq:fea_div}) to optimally select a mesh tile containing a variety of classes and objects at different scales and complexity.
The feature diversity $F_{m}$ of tile $m$ is computed as
\begin{equation}\label{eq:fea_div}
	F_{m}=\frac{\sum_{i=1}^{N_{f}}\left ( f_i - \bar{f} \right )^{2}}{N_{f}}
\end{equation}
where $f_i$ represents each handcrafted feature which describe in Section \ref{sec:initial_seg}, and $\bar{f}$ is mean value of a $N_{f}$ dimensional feature vector.
To acquire the first ground truth data, we manually annotate the mesh (with segments) that is selected with the highest feature diversity.
Then, we add the first labelled mesh into the training dataset for the supervised classification.
Specifically, we use the segment-based features as input for the classifier, and the output is a pre-labelled mesh dataset.
Next, we use the mesh annotation tool to manually refine the pre-labelled mesh according to the feature diversity.
Finally, the new refined mesh will be added to the training dataset to improve the automatic classification accuracy incrementally.

\subsubsection{Initial Segmentation}\label{sec:initial_seg}

To avoid redundant computations of numerous triangles, we first apply mesh over-segmentation (i.e., linear least-squares fitting of planes) based on region growing on the input data to group triangle faces into homogeneous regions~\citep{lafarge2012creating}.
Such grouped regions are beneficial for computing local contextual features.
We then extract both geometric and radiometric features from those mesh segments as follows: 
\begin{itemize}
	\item[$\bullet$] \textit{Eigen-based features} are computed from the covariance matrix of the triangle vertices with respect to the average centre within each segment, which is beneficial for identifying urban objects with various surface distributions.
	The linearity $= (\lambda_{1} - \lambda_{2}) / \lambda_{1}$, sphericity $= \lambda_{3}/ \lambda_{1}$ and change of curvature $= \lambda_{3} / (\lambda_{1} + \lambda_{2} + \lambda_{3})$ are computed based on the three eigenvalues $\lambda_{1} \geq \lambda_{2} \geq \lambda_{3}\geq 0$.
	The local eigenvectors $\mathbf{n}_{i} $ and the unit normal vector $\mathbf{n}_{z} $ along Z-axis are used to compute the verticality $=1-\left | \mathbf{n}_{i}\cdot \mathbf{n}_{z} \right | $~\citep{hackel2016fast}.
	Note that many eigen-based features have been studied in literature~\citep{hackel2016fast,west2004context,weinmann2013feature}, and some of them were designed for and tested on LiDAR point clouds. 
	\textcolor{ao}{
	These eigen-based features are mostly computed per point based on its spherical neighbourhood, which often contains noise and does not form a surface. 
	Our chosen eigen-based features are defined on a segment representing the surface of a mesh, and thus they can capture non-local geometric properties of an object.
	}
	Additionally, in this work, we have tested all eigen-based features from the literature~\citep{hackel2016fast}, and we only present the ones that are effective for texture meshes.
	\item[$\bullet$] \textit{Elevation} is divided into absolute elevation $z_{a}$, relative elevation $z_{r}$ and multiscale elevations $z_{m}$.
	Where $z_{a}$ is the average elevation of the segment;
	the relative elevation is computed as $z_{r} = z_{a}-z_{r_{min}}$;
	the multiscale elevation~\citep{Verdie2015,Rouhani2017} $z_{m} = \sqrt{\frac{z_{a} - z_{min}}{z_{max} - z_{min}}}$.
	And $z_{r_{min}}$ denotes the lowest elevation of the local largest ground segment computed within a cylindrical neighbourhood with 30 meters radius around the segment centre.
	$z_{min}$ and $z_{max}$ represent the local minimum and maximum elevation values of a cylindrical neighbourhood within the scale of 10 meters, 20 meters, and 40 meters.
	Such large cylindrical neighbourhoods allow to find the local ground considering the resilience to hilly environments, \textcolor{ao}{and the square root ensures that small relative height values (i.e., values smaller than 1 $ m $) get a larger elevation attribute to enlarge elevation differences between small objects and the local ground (e.g., cars against the ground, boats against the water surfaces).}
	More importantly, due to the influence of terrain fluctuations and various scales of urban objects, the elevation of these three categories can complement each other.
	\item[$\bullet$] \textit{Segment area} is computed as $area(S_k) = \sum_{i = 1}^{N} area(f_i) $, where $f_i$ denotes a triangle of the segment $S_k$, and $N$ denotes the total number of triangles in $S_k$.
	\item[$\bullet$] \textit{Triangle density} is defined as $density(S_k) = \frac{N}{area(S_k)} $,  which reveals the object complexity, especially for adaptive urban meshes.
	\item[$\bullet$] \textit{Interior radius of 3D medial axis transform (InMAT)}~\citep{ma20123d,peters2016robust} of a segment $S_k$ is formulated as $r_k = \frac{\sum_{i=1}^{M} r_i}{M}$, where $M$ denotes the total number of triangle vertices of $S_k$, and $r_i$ denotes the interior radius of the shrinking ball that touches the vertex $v_i$ within the segment $S_k$. 
	It is designed to distinguish objects with different scales. 
	\item[$\bullet$] \textit{HSV colour-based features} are derived from the RGB channel of the entire texture map.
	We use the HSV colour space since it can better differentiate different objects than RGB.
	We compute the average colour, the variance of the colour distribution of all pixels within each segment, and we further discretize it into a histogram that consists of 15 bins of the hue channel, five bins of the saturation channel, and five bins of the value channel.
	\item[$\bullet$] \textit{Greenness} $a_{g}$ is used to classify objects that are similar to green vegetation.
	Specifically, it is computed according to the averaged RGB colour of each segment via $a_{g}=G-0.39\cdot R-0.61\cdot B$~\citep{mckinnon2017comparing}. 
\end{itemize}
	All the above features are concatenated into a 44-dimensional feature vector used by our random forest (RF) classifier in the initial segmentation. 

\subsubsection{Annotation Tool for Refinement}

\begin{figure}[!tb]
	\centering
	\includegraphics[width=\textwidth]{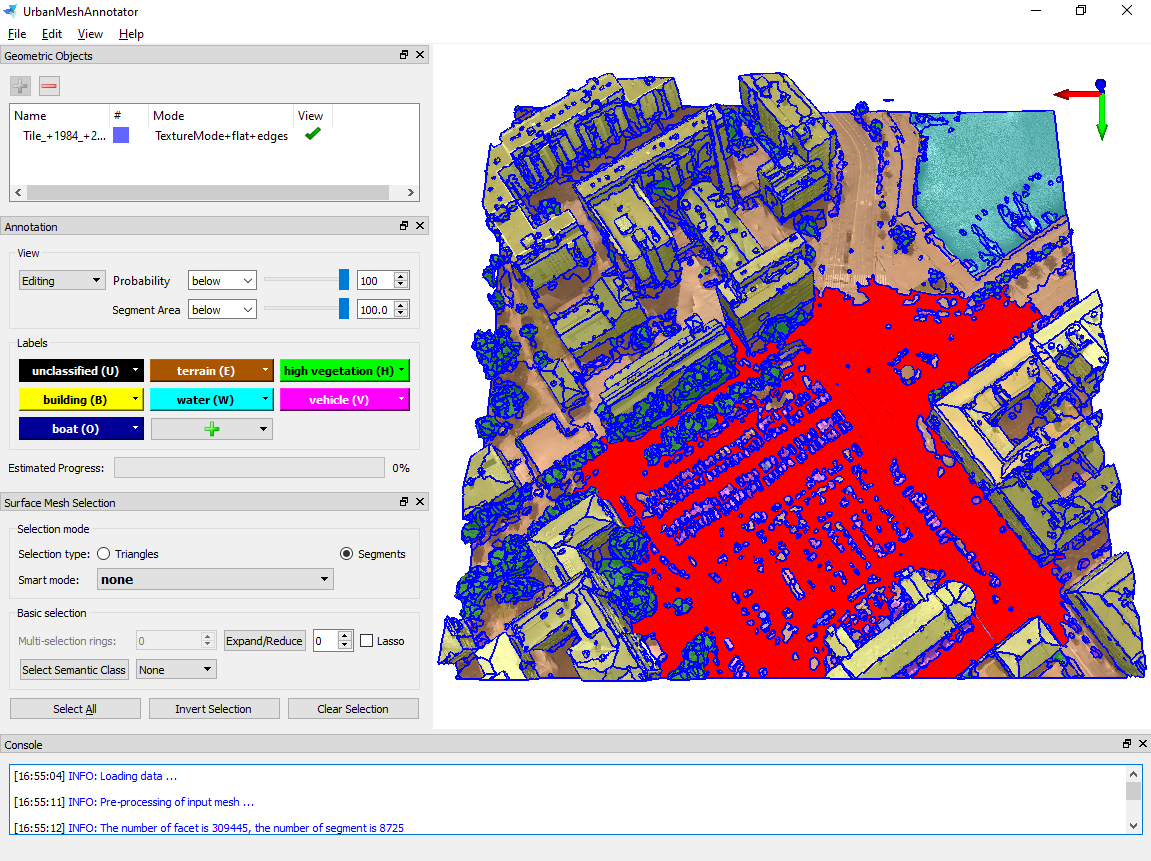}
	\caption{The interface of our annotation tool for 3D texture meshes. }
	\label{fig:annotator}
\end{figure}

Because of the under-segmentation errors and the imperfect results of the semantic mesh segmentation process, we design a mesh annotation tool (see Figure \ref{fig:annotator}) to manually correct the labelling errors.
Our mesh annotation tool is developed based on the labelling tool of CGAL~\citep{cgal:eb-20b}.

As shown in Table \ref{tab:annotation_operation}, it consists of three operation categories: view, selection, and annotation.
The	view operations provide essential functions for the user to manipulate the scene camera, such as translate, rotate, zoom, or set the new pivot for the scene.
In addition, to use textures as a reference for labelling, we map texture and face colour with a certain degree of transparency, and we visualize the segment border to differentiate each segment. 

\begin{table}[!tb]
	\centering
	\noindent\adjustbox{max width=0.8\textwidth}
	{
		\begin{threeparttable}
			\centering
			\begin{tabular}{ccc}
				\toprule
				Categories & Operations & Objects \\
				\midrule
				\multirow{4}[2]{*}{View} & Translate & Camera \\
				& Rotate & Camera \\
				& Zoom in / out & Camera \\
				& Set pivot & Camera \\
				\midrule
				\multirow{6}[2]{*}{Selection} & Multi-selection / Lasso & Triangles / Segments \\
				& Expand / Reduce & Triangles / Segments \\
				& Semantic selection & Segments \\
				& Split region & Segments \\
				& Planar region extraction & Triangles \\
				& Split mesh & Triangles \\
				\midrule
				\multirow{3}[2]{*}{Annotation} & Probability slider & Segments \\
				& Segment area slider & Segments \\
				& Progress bar & Triangles \\
				& Switch semantic view & Triangles \\ 
				& Labelling & Triangles / Segments \\
				\bottomrule
			\end{tabular}%
		\end{threeparttable}
	}
	\caption{Basic operations in our annotation tool.} 
	\label{tab:annotation_operation}%
\end{table}%

\begin{figure}[!tb]
	\centering
	\begin{subfigure}[t]{0.32\textwidth}
		\includegraphics[width=\linewidth]{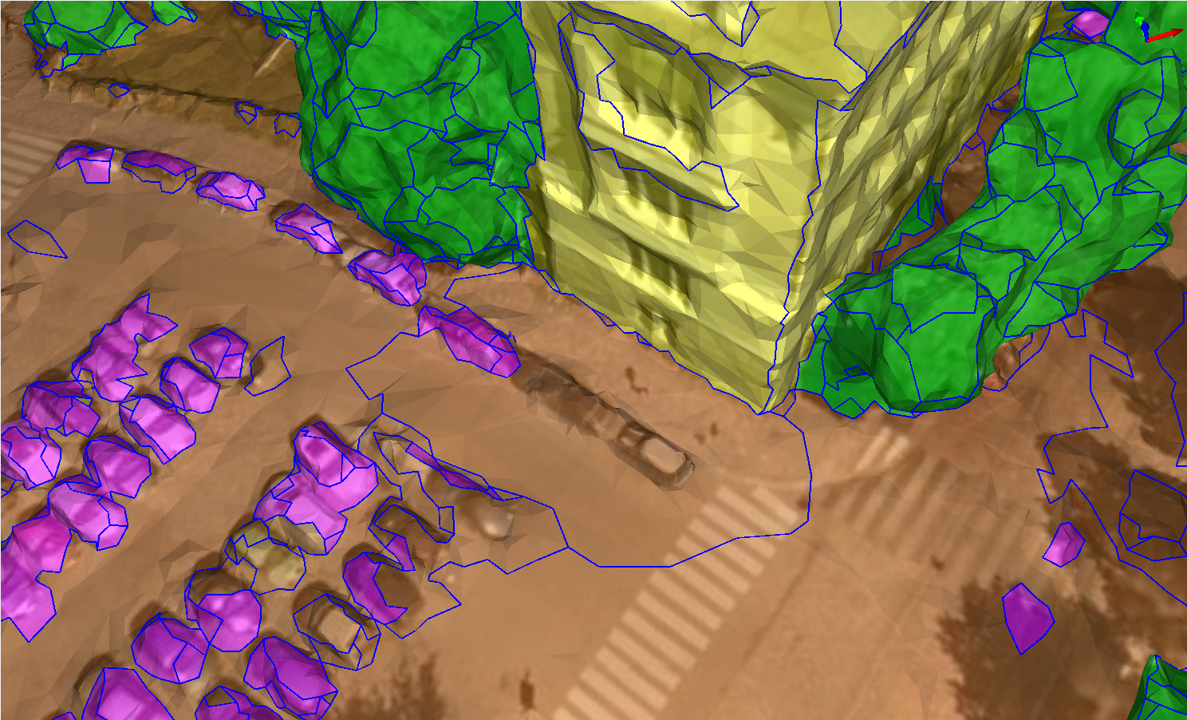}
		\caption{}
	\end{subfigure}
	\hspace*{\fill}
	\begin{subfigure}[t]{0.32\textwidth}
		\includegraphics[width=\linewidth]{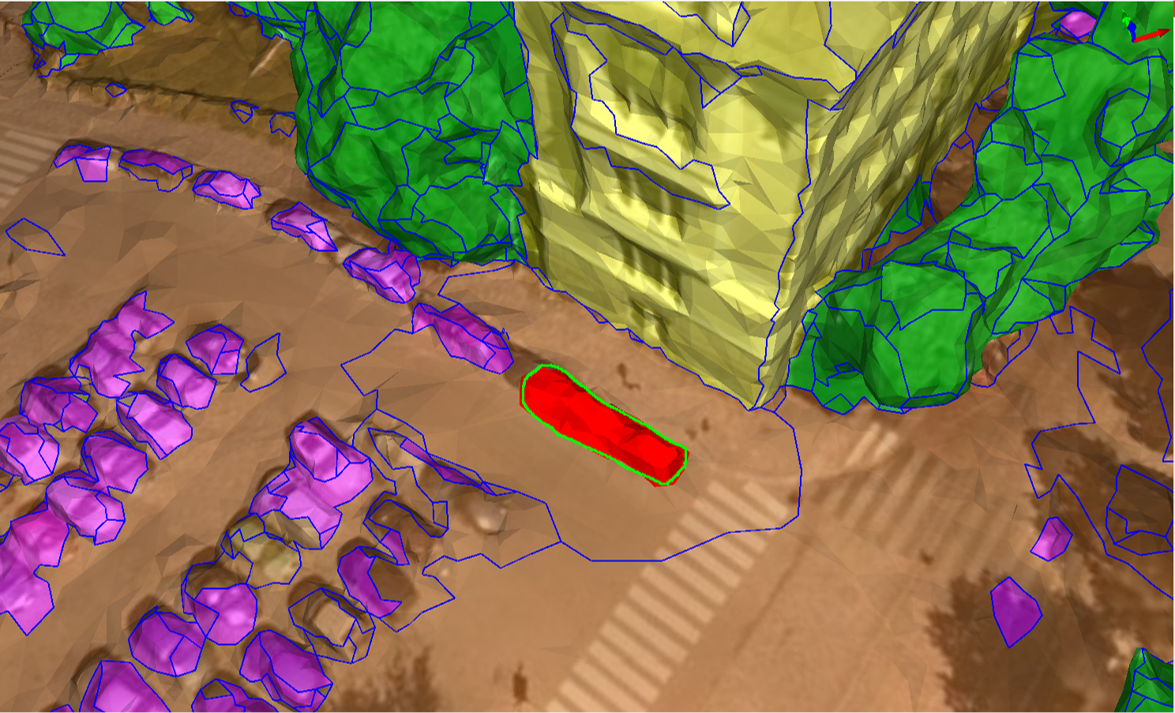}
		\caption{}
	\end{subfigure}
	\hspace*{\fill}
	\begin{subfigure}[t]{0.32\textwidth}
		\includegraphics[width=\linewidth]{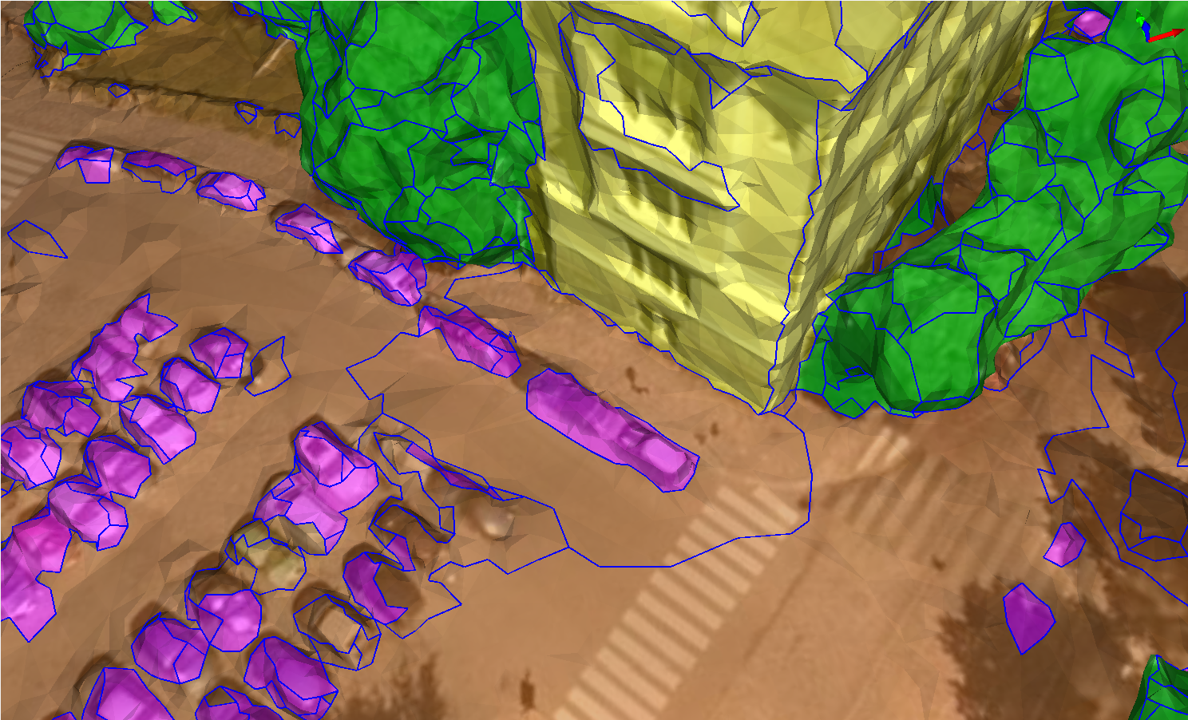}
		\caption{}
	\end{subfigure}
	\caption{An example of labelling by selecting triangles using the lasso tool (blue edges: segment boundaries). 
		(a) Before selection.
		(b) Lasso selection result (in red).
		(c) The correct label has been assigned to the selected region. 
		In this example, the label of the selected region has been changed from `ground' to `vehicle'.
	} 
	\label{fig:tri_sel}
\end{figure}

\begin{figure}[!tb]
	\centering
	\begin{subfigure}[t]{0.32\textwidth}
		\includegraphics[width=\linewidth]{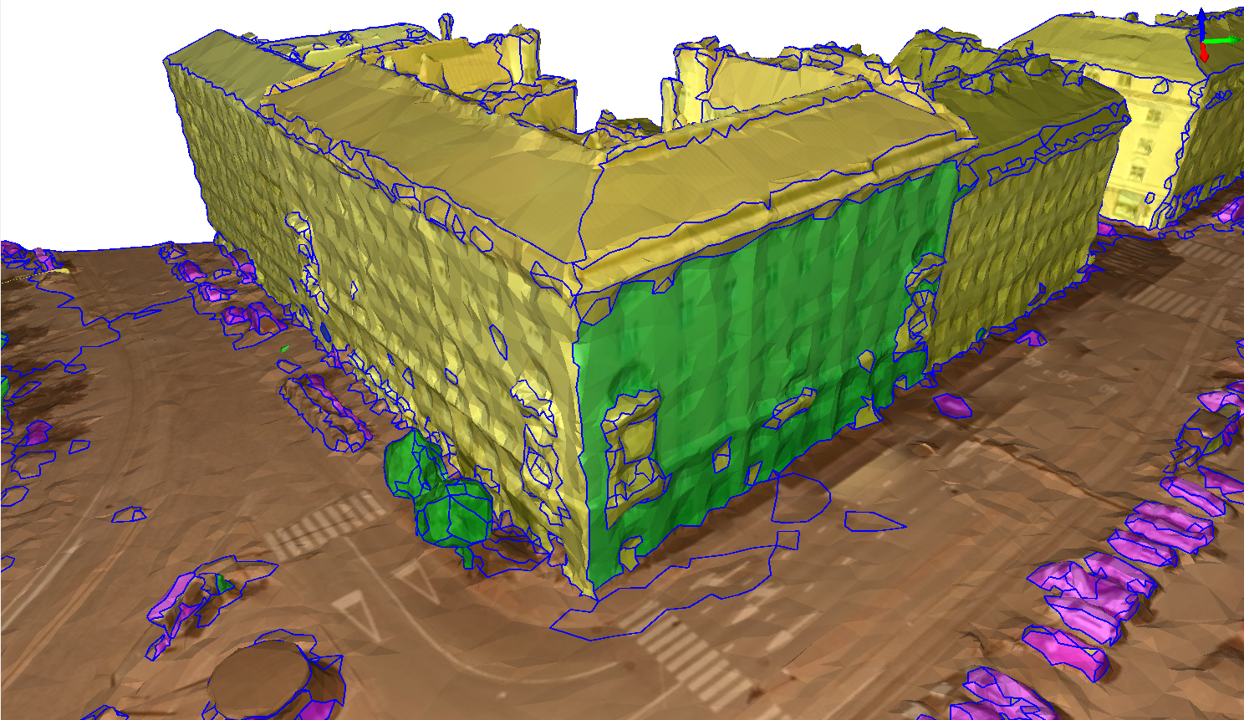}
		\caption{}
	\end{subfigure}
	\hspace*{\fill}
	\begin{subfigure}[t]{0.32\textwidth}
		\includegraphics[width=\linewidth]{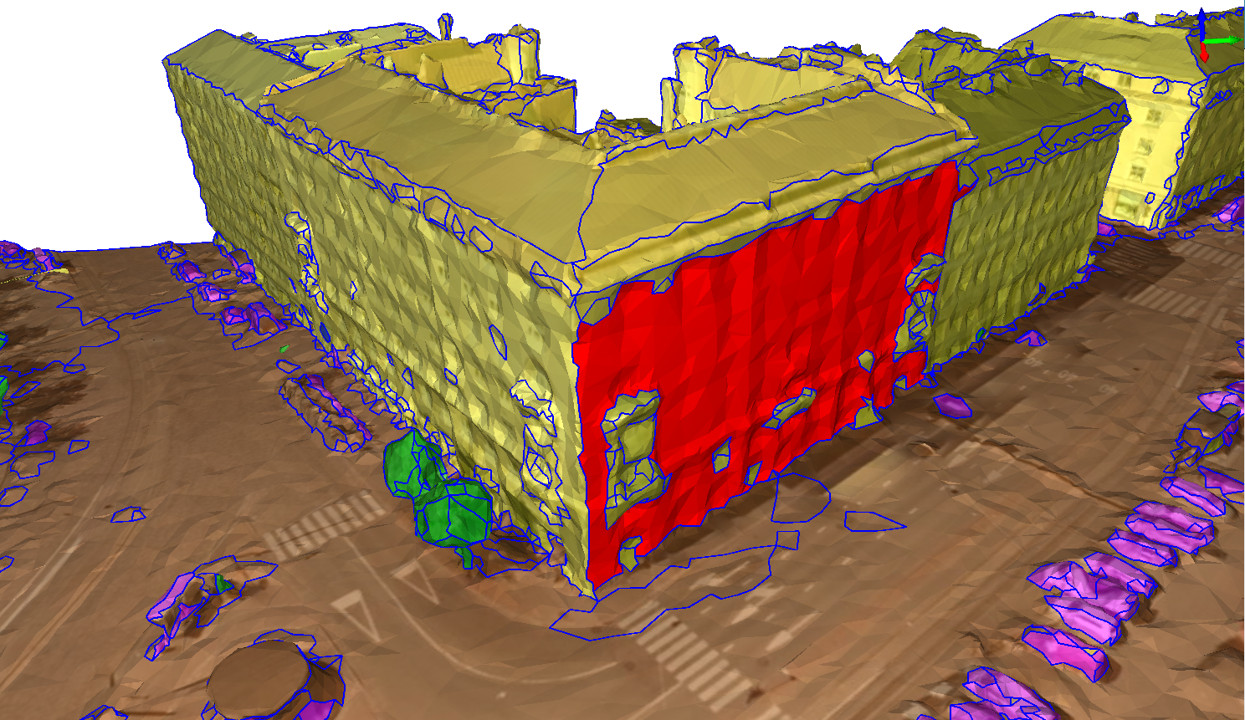}
		\caption{}
	\end{subfigure}
	\hspace*{\fill}
	\begin{subfigure}[t]{0.32\textwidth}
		\includegraphics[width=\linewidth]{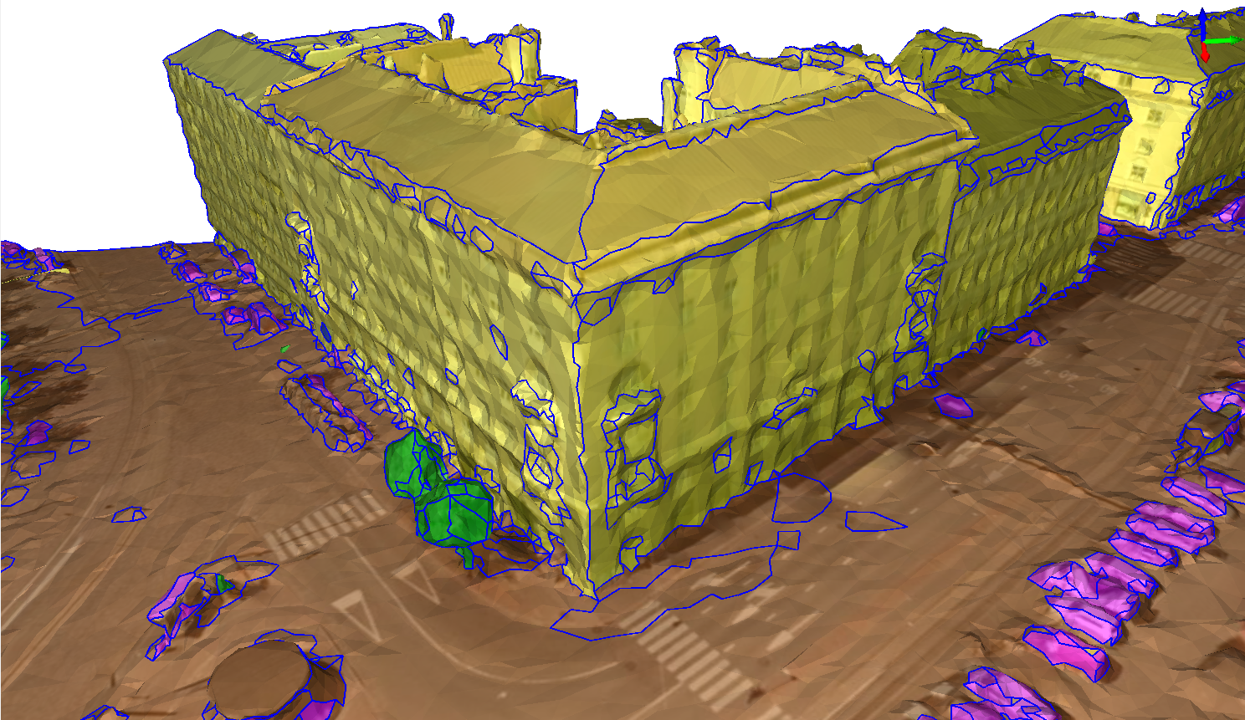}
		\caption{}
	\end{subfigure}
	\caption{An example of segment labelling. 
		(a) Part of a wall of the building was previously labelled as `high vegetation' (in green).
		(b) Segment selection result (in red).
		(c) The label of the selected segment has been corrected with the new label `building'.
	}
	\label{fig:seg_sel}
\end{figure}

The	selection operations allow the user to select or deselect either triangle faces (see Figure \ref{fig:tri_sel}) or segments (see Figure \ref{fig:seg_sel}) freely via a brush or a lasso.
Specifically, the face selection operation is used to fix the under-segmentation errors and generate new segments, and the segment selection operation is to fix incorrect segment labels.

\begin{figure}[!tb]
	\centering
	\begin{subfigure}[t]{0.48\textwidth}
		\includegraphics[width=\linewidth]{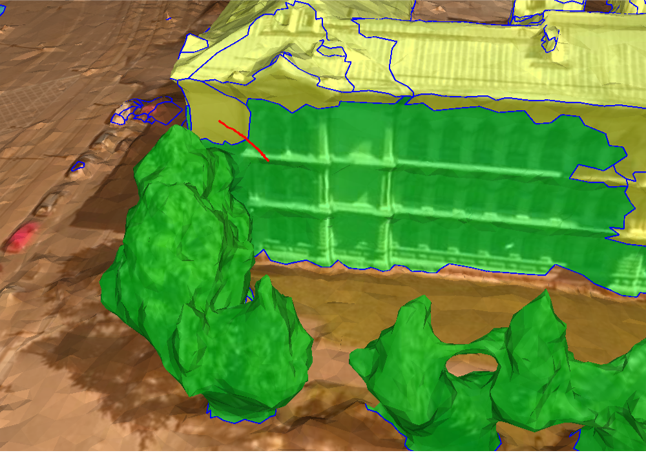}
		\caption{}
	\end{subfigure}
	\hspace*{\fill}
	\begin{subfigure}[t]{0.48\textwidth}
		\includegraphics[width=\linewidth]{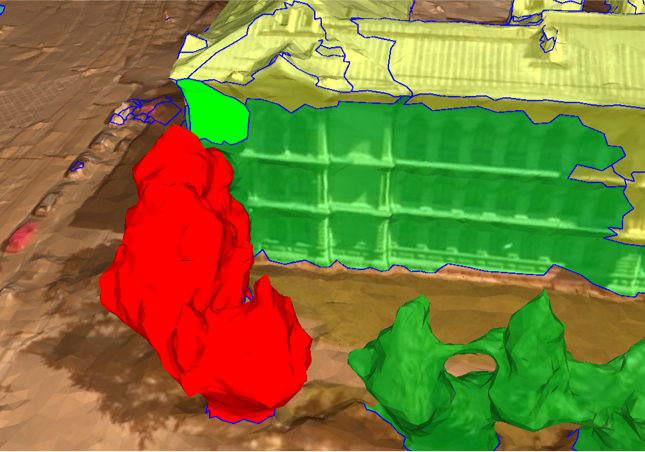}
		\caption{}
	\end{subfigure}
	\caption{An example splitting planar and non-planar regions. 
		(a) The user draws a stroke (in red) across the border of the non-planar segment and the planar segment. 
		(b) The detected non-planar segment has been split into two parts (i.e., a non-planar region shown in red and a planar segment shown in green).
	} 
	\label{fig:pnp_func}
\end{figure}

\begin{figure}[!tb]
	\centering
	\begin{subfigure}[t]{0.48\textwidth}
		\includegraphics[width=\linewidth]{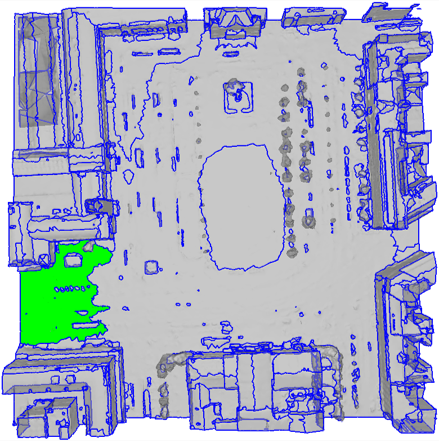}
		\caption{}
	\end{subfigure}
	\hspace*{\fill}
	\begin{subfigure}[t]{0.48\textwidth}
		\includegraphics[width=\linewidth]{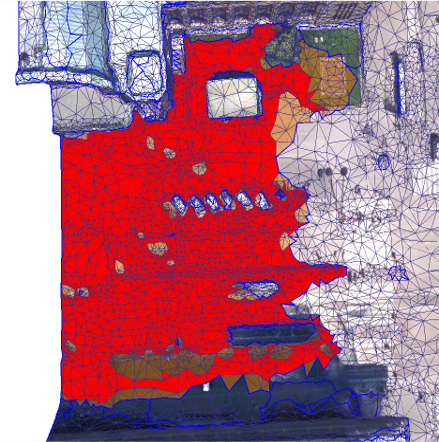}
		\caption{}
	\end{subfigure}
	\caption{Editing an individual segment. 
		(a) A segment is selected (highlighted in green) for splitting. 
		(b) Automatic extraction of the most planar region (shown in red) within the selected segment according to user-defined thresholds.} 
	\label{fig:seg_func}
\end{figure}

We also allow the user to edit the selection of each individual segment with splitting functions (see Figure \ref{fig:pnp_func}) and automatic extraction of the most planar region (see Figure \ref{fig:seg_func}). 
As for splitting, we first detect the potential planar and non-planar segments marked by user strokes, and then the non-planar one is split according to the vertex-to-plane distance.
It allows generating candidate non-planar regions (with respect to the detected planar segment) for the user to edit, and
it is useful to split a segment that covers large non-planar regions or contains more than one dominant planar area.
To extract the most planar region, we apply the region growing algorithm~\citep{lafarge2012creating} within the selected segment to automatically generate the candidate triangle faces with user-defined thresholds (i.e., the maximum distance to the plane, the maximum accepted angle, and the minimum region size).
Such an operation allows the user to filter out some small bumpy regions of the selected segment.

Besides, probability and area-based sliders and a progress bar are provided in the annotation panel to improve annotation efficiency and experience, respectively. 
Specifically, the probability slider is introduced for the user to visually inspect the segments that are most likely misclassified.
Moreover, the user can further use it to inspect a specific class by switching the view to highlight a specific semantic class.
The segment area slider is used to identify isolated tiny segments, which commonly appear as errors.
The progress bar is used to indicate the estimated labelling progress during the annotation.
After performing the selection, the user can easily assign the corresponding label to the selected area.

	\section{Experiments}\label{sec:expri}
\subsection{Data Split}

\begin{figure}[!tb]
	\begin{subfigure}[b]{0.38\textwidth}
		\includegraphics[width=\linewidth]{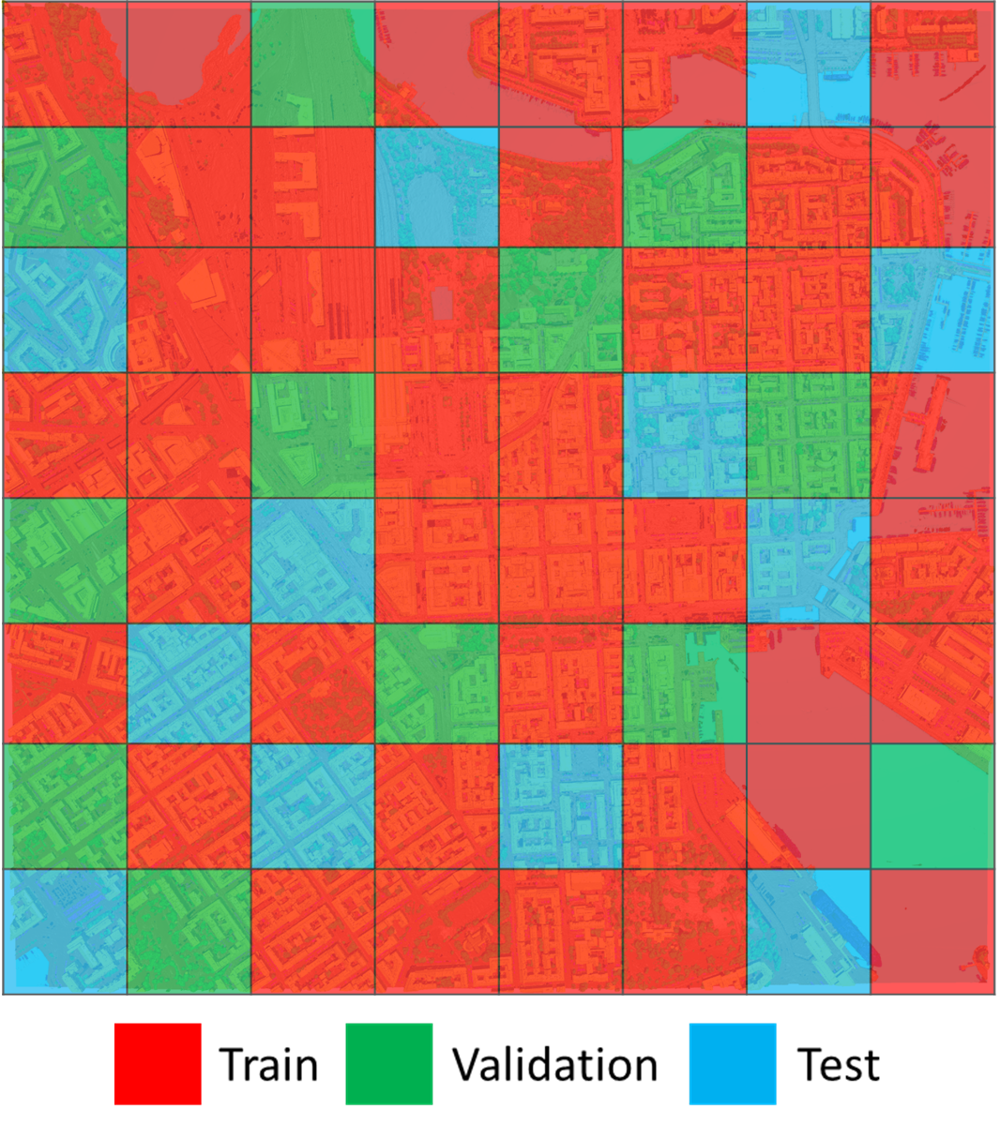}
		\label{fig:grids}
		\caption{}
	\end{subfigure}
	\hspace*{\fill}
	\begin{subfigure}[b]{0.58\textwidth}		
		\includegraphics[width=\linewidth]{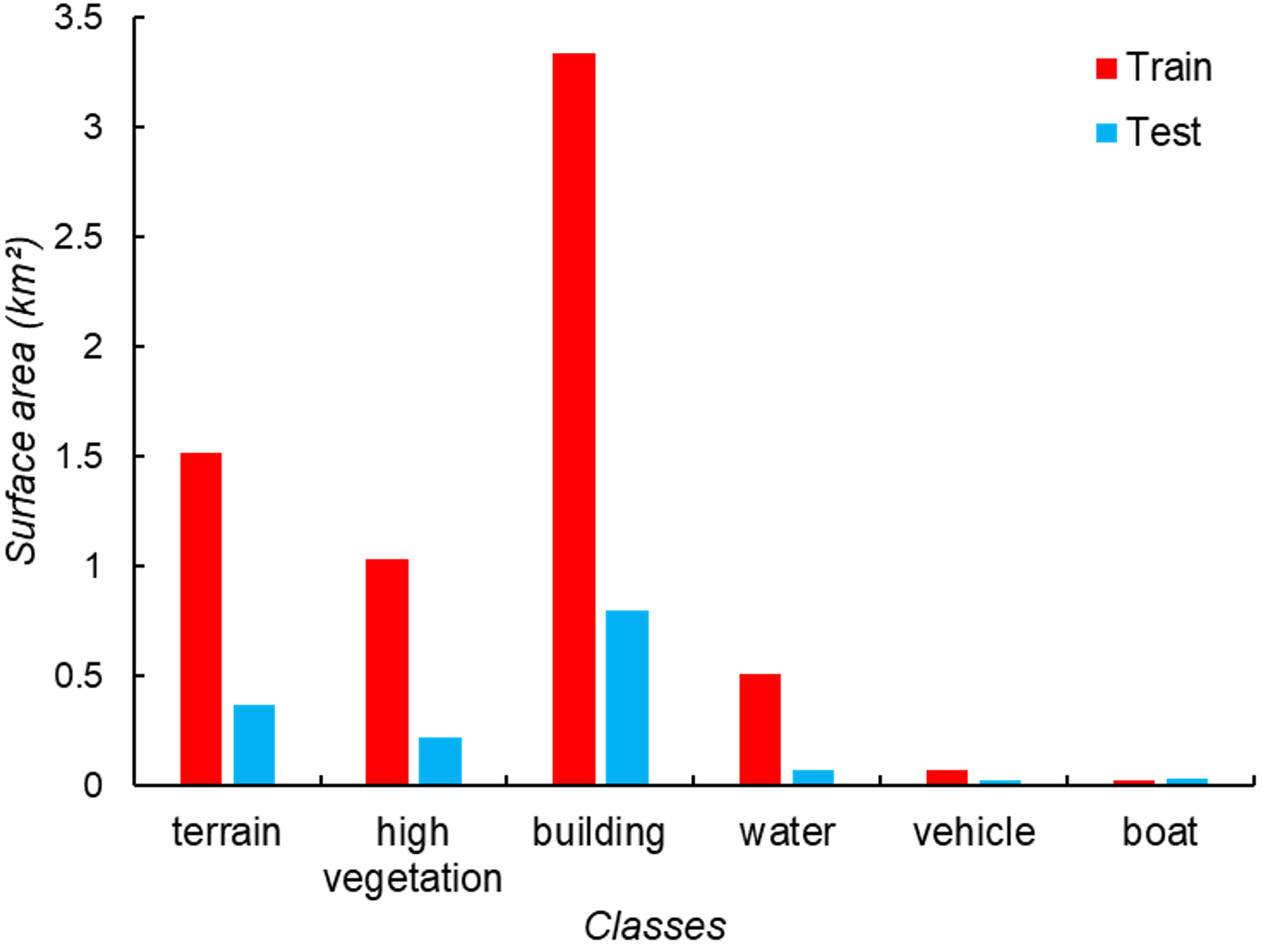}
		\label{fig:class_statistics}
		\caption{}
	\end{subfigure}
	\caption{Overview of the data used in our experiment. (a) The distribution of the training, test, and validation dataset. (b) Semantic categories of training (including validation dataset) and test dataset.}
	\label{fig:datadesp}
\end{figure}

To perform the semantic segmentation task, we randomly select 40 tiles from the annotated 64 tiles of Helsinki as training data, 12 tiles as test data, and 12 tiles as validation data (see Figure \ref{fig:datadesp} (a)).
For each of the six semantic categories, we compute the total area in the training and test dataset to show the class distribution.
As shown in Figure \ref{fig:datadesp} (b), some classes, like vehicles and boats, only account for less than $5\%$ of the total area,
while the building and terrain together comprise more than $70\%$.
The unbalanced classes impose significant challenges for semantic segmentation based on supervised learning.

\subsection{Evaluation Metric}
Since the triangle faces in the meshes have different sizes, we compute the surface area for semantic evaluation instead of using the number of triangles. The performance of semantic mesh segmentation is measured in precision, recall, F1 score, and intersection over union (IoU) for each object class. The evaluation of the whole test area is applied with overall accuracy (OA), mean per-class accuracy (mAcc), and mean per-class intersection over union (mIoU).

\subsection{Evaluation of Initial Segmentation}
We have implemented the semantic mesh segmentation and annotation tool in C++ using the open-source libraries include CGAL~\citep{cgal:eb-20b}, Easy3D~\citep{easy3d2018nan}, and ETHZ random forest~\citep{ethzrf}. 

Our proposed pipeline for initial segmentation only takes a few input parameters, which are shown in Table \ref{tab:params}.
The over-segmentation is intended to find all planar regions in the model, for which we set the distance threshold to 0.5 meters. This threshold value specifies the minimum geometric features we would like the over-segmentation method to identify. 
In other words, the region growing-based over-segmentation method will not be able to distinguish two parallel planes with a distance smaller than this threshold.
We set the angle threshold to 90 degrees, which is large enough to cope with high levels of noise (e.g., the distance value is small, but the angle between the triangle normal and the plane normal is large).
Moreover, the minimum area is set to zero to allow planar segments of any arbitrary size.
As for the random forest classifier, \textcolor{ao}{we set the parameters initially to those of Rouhani et al.~\citep{Rouhani2017} followed by fine-tuning using the validation data.}
Specifically, using 100 trees is sufficient to guarantee the stability of the model, and using the depth of 30 is adequate to avoid over-fitting and under-fitting for training.
\begin{table}[H]
	\centering
	\noindent\adjustbox{max width=0.6\textwidth}
	{
		\begin{threeparttable}
				\centering
				\begin{tabular}{ccc}
					\toprule
					Method & Parameters & Value \\
					\midrule
					\multirow{3}[2]{*}{Region Growing} & Minimum area & 0 $m^2$ \\
					& Distance to plane & 0.5 \textcolor{ao}{$m$} \\
					& Accepted angle & $90^{\circ}$ \\
					\midrule
					\multirow{2}[2]{*}{Random Forest} & Number of trees & 100 \\
					& Maximum depth & 30 \\
					\bottomrule
				\end{tabular}%
		\end{threeparttable}
	}
	\caption{Parameters used in our approach.} 
	\label{tab:params}
\end{table}%

Rather than classifying about 19 million triangle faces (i.e., the entire dataset), we use 515,176 segments that are clustered during over-segmentation.
Although both semantic segmentation and labelling refinement can benefit from mesh over-segmentation, the degree of the under-segmentation error cannot be avoided.
Since our mesh over-segmentation does not intend to retrieve the individual objects and the purpose is to perform semantic segmentation, we measure the maximum achievable performance by calculating the \textcolor{ao}{IoU} instead of using under-segmentation errors to evaluate it.
The \textcolor{ao}{upper bound IoU} of each class we could achieve for semantic segmentation is presented in Table \ref{tab:oveal_re}, and the \textcolor{ao}{upper bound mean IoU (mIoU)} over all classes is about $ 90.9\% $ as shown in Table \ref{tab:fea_abla}.
In addition, the results of our experiment in Tables \ref{tab:oveal_re} and \ref{tab:fea_abla} are reported based on the average performance of ten times experiments with the same configuration.

\begin{table}[H]
	\centering
	\noindent\adjustbox{max width=1.0\textwidth}
	{
		\begin{threeparttable}
			\centering
			\begin{tabular}{cccccC{0.15\linewidth}}
				\toprule
				Class & Precision  (\%) & Recall  (\%) & F1 scores  (\%) & IoU  (\%) & Upper bound IoU  (\%) \\
				\midrule
				Terrain & 87.7 & 94.3 & 90.9 & 83.3 & 93.9 \\
				High Vegetation & 96.3 & 93.8 & 95.0 & 90.5 & 96.2 \\
				Building & 94.6 & 97.7 & 96.1 & 92.5 & 99.0 \\
				Water & 97.0 & 88.3 & 92.5 & 86.0 & 92.7 \\
				Vehicle & 77.9 & 41.7 & 54.4 & 37.3 & 73.2 \\
				Boat & 77.9 & 7.5 & 13.7 & 7.4 & 90.5 \\
				\bottomrule
			\end{tabular}%
		\end{threeparttable}
	}
	\caption{Overall evaluation of our method.
	The \textit{Upper bound IoU} refers to the maximum achievable IoU in theory.}
	\label{tab:oveal_re}
\end{table}%

\begin{figure}[H]
	\begin{adjustwidth}{-3cm}{-3cm}
		\centering
		\begin{tabular}{ccccc}	
			\includegraphics[width=0.25\textwidth]{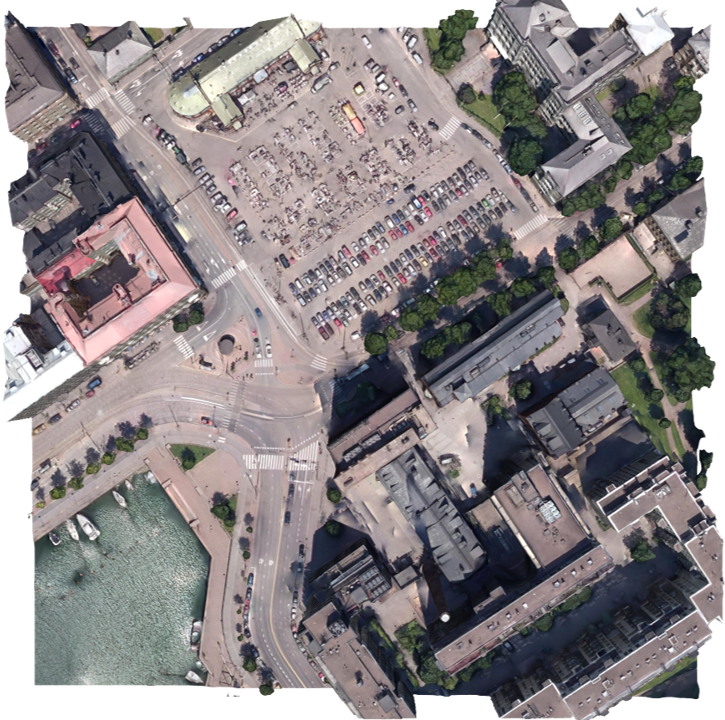}&
			\includegraphics[width=0.25\textwidth]{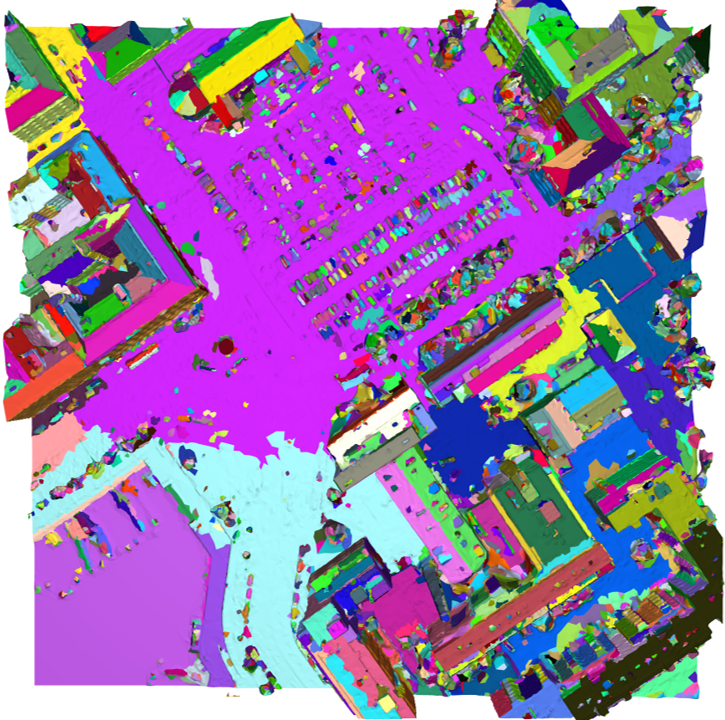}&
			\includegraphics[width=0.25\textwidth]{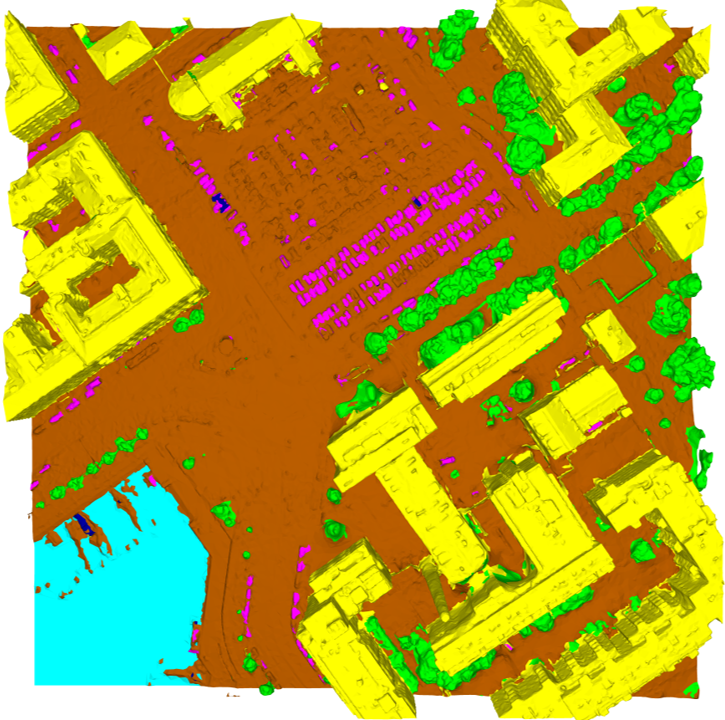}&
			\includegraphics[width=0.25\textwidth]{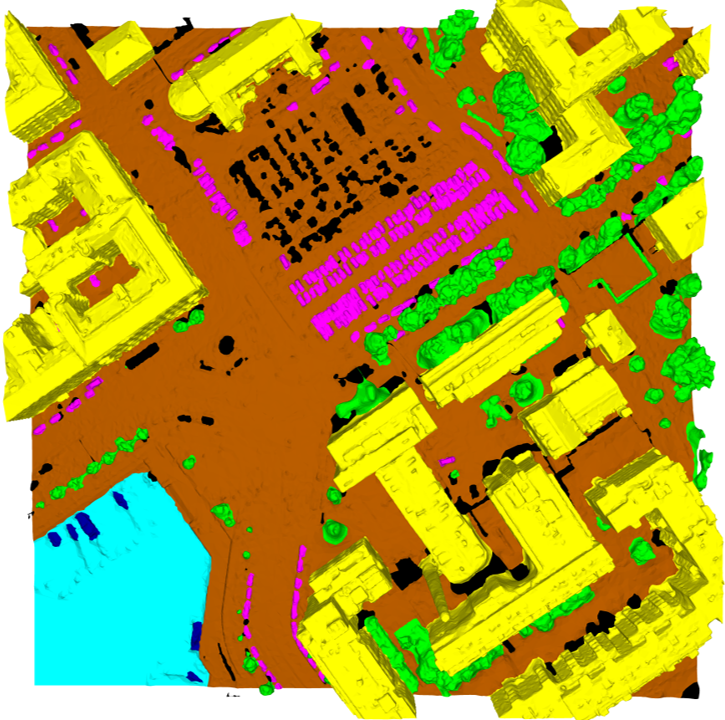}&
			\includegraphics[width=0.25\textwidth]{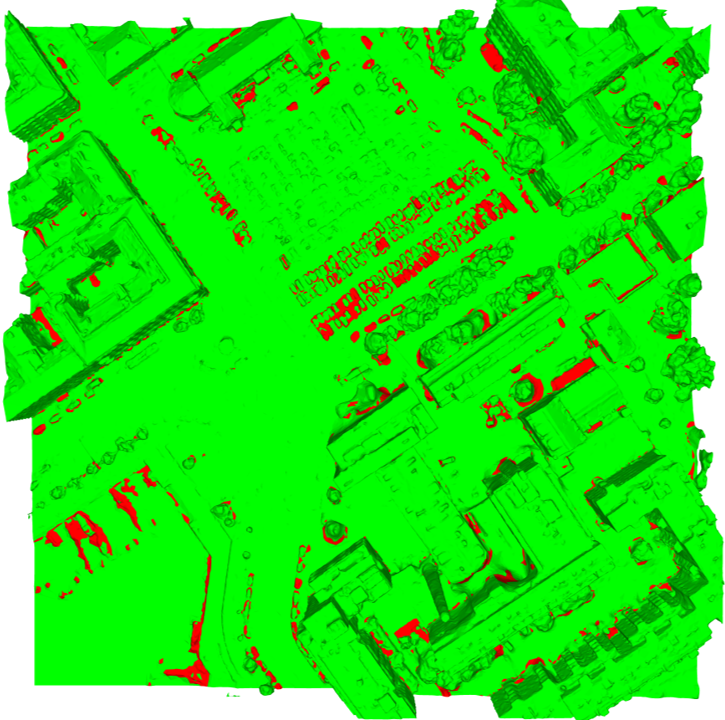}\\
			\midrule
			\includegraphics[width=0.25\textwidth]{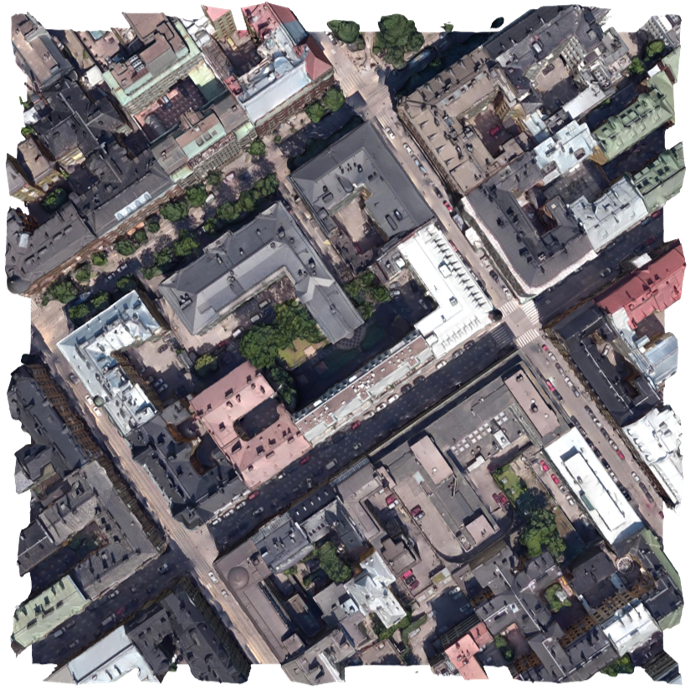}&
			\includegraphics[width=0.25\textwidth]{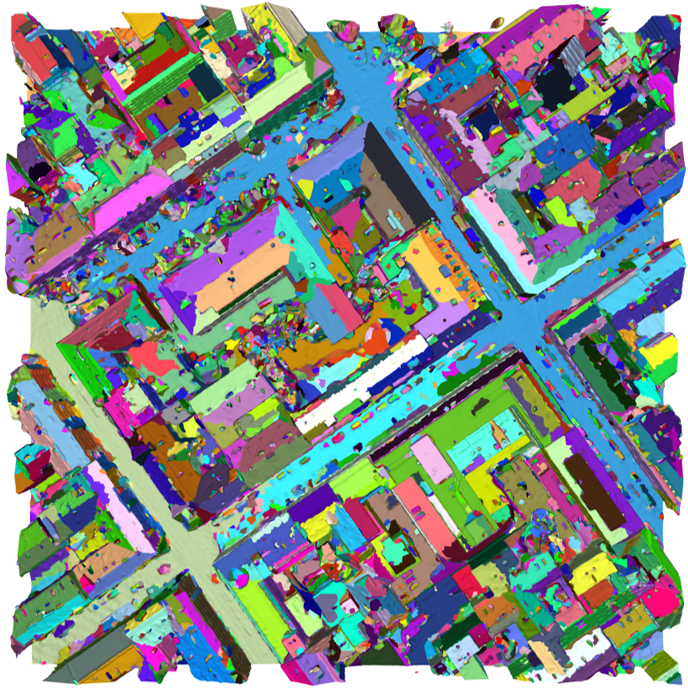}&
			\includegraphics[width=0.25\textwidth]{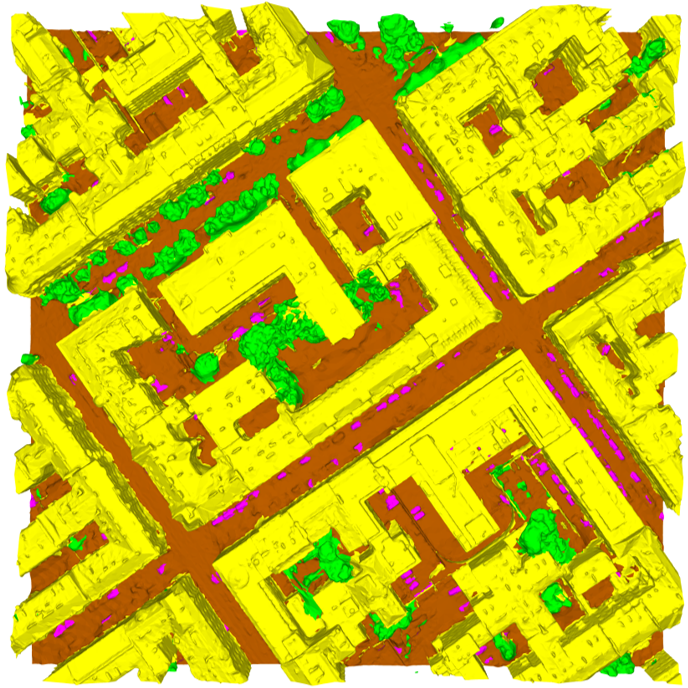}&
			\includegraphics[width=0.25\textwidth]{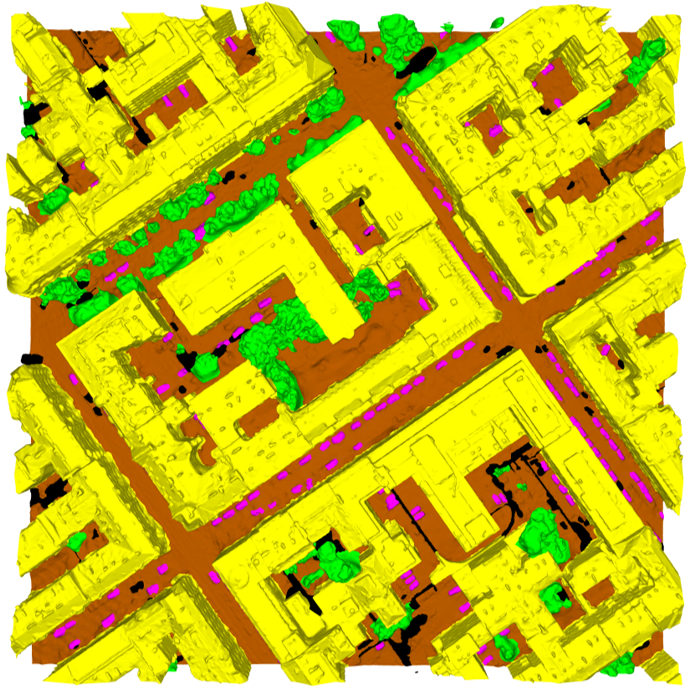}&
			\includegraphics[width=0.25\textwidth]{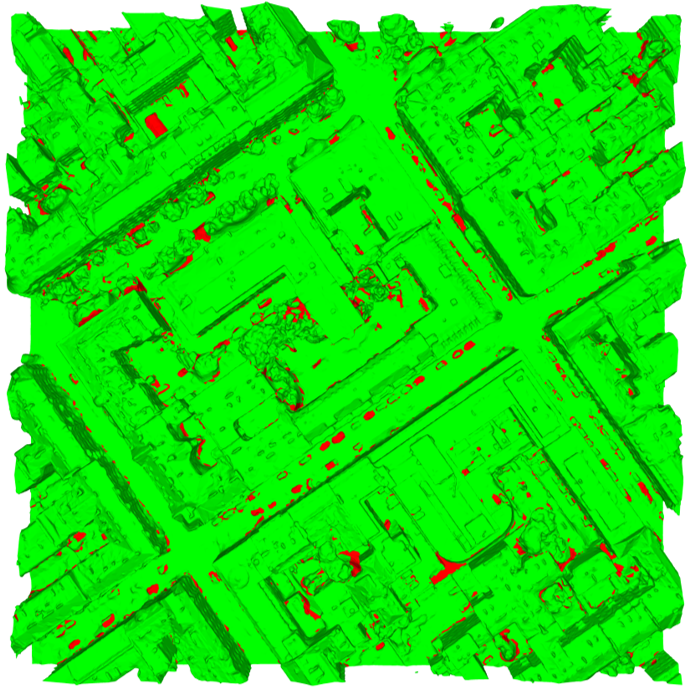}\\
			\midrule
			\includegraphics[width=0.25\textwidth]{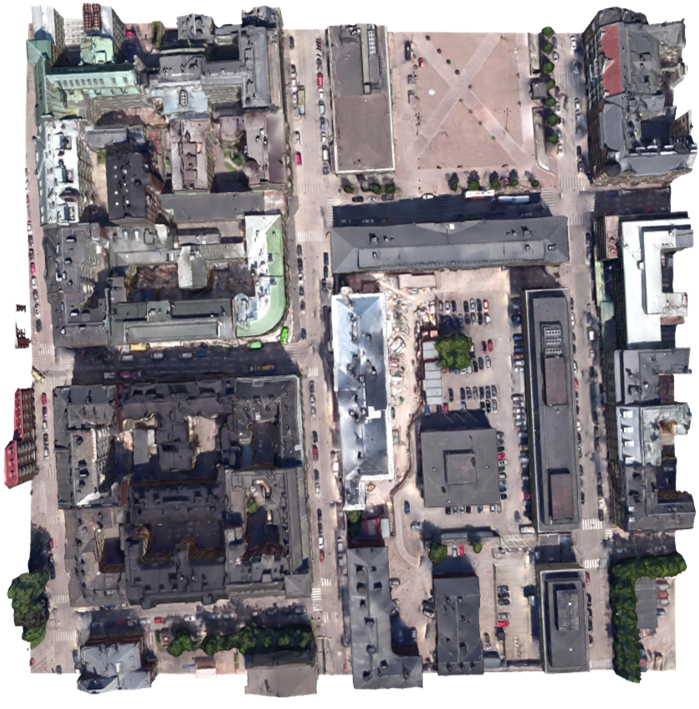}&
			\includegraphics[width=0.25\textwidth]{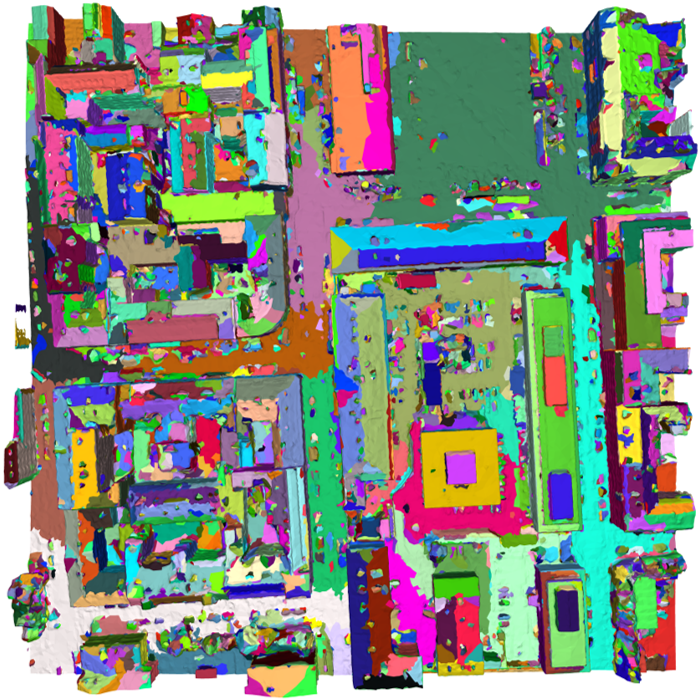}&
			\includegraphics[width=0.25\textwidth]{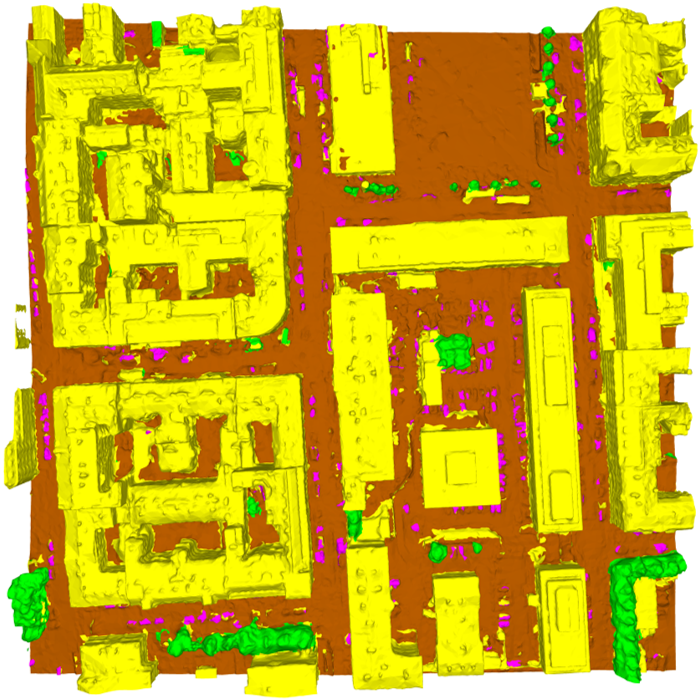}&
			\includegraphics[width=0.25\textwidth]{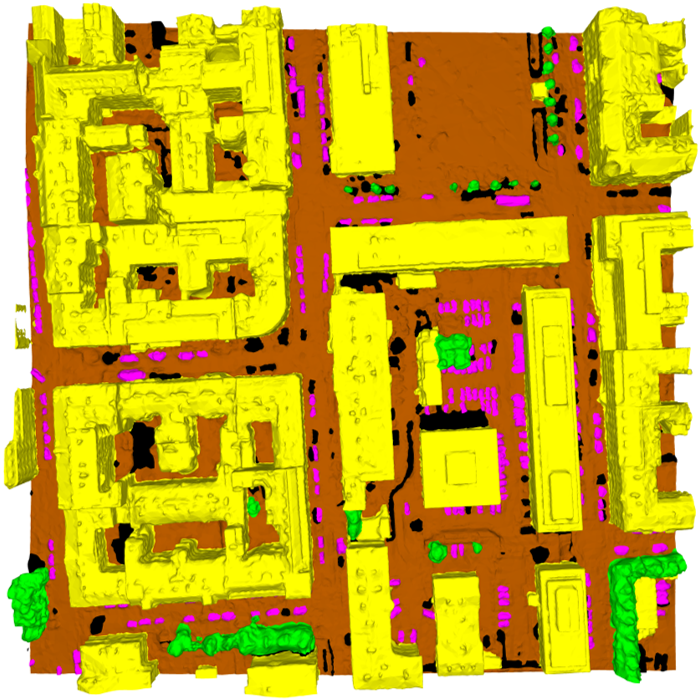}&
			\includegraphics[width=0.25\textwidth]{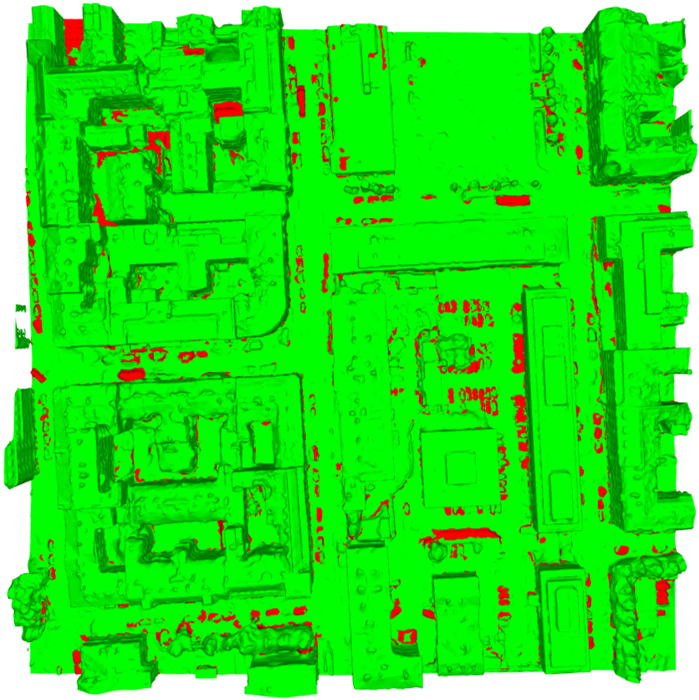}\\
			\midrule
			\includegraphics[width=0.25\textwidth]{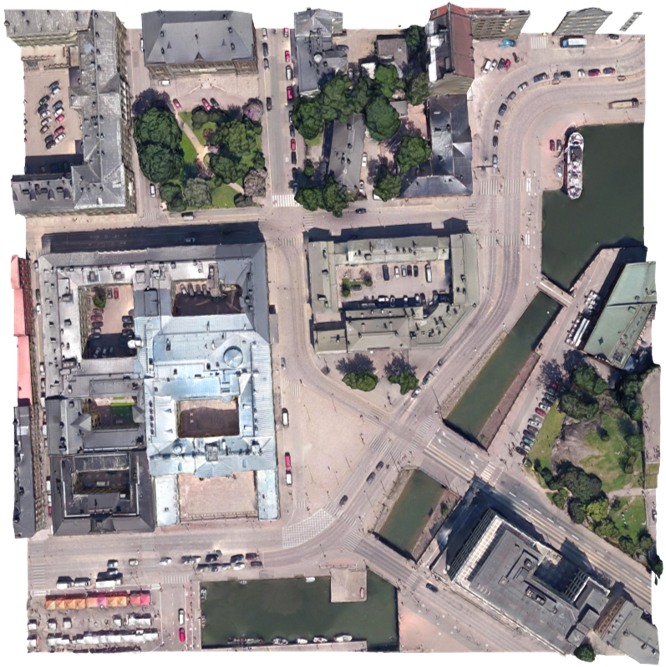}&
			\includegraphics[width=0.25\textwidth]{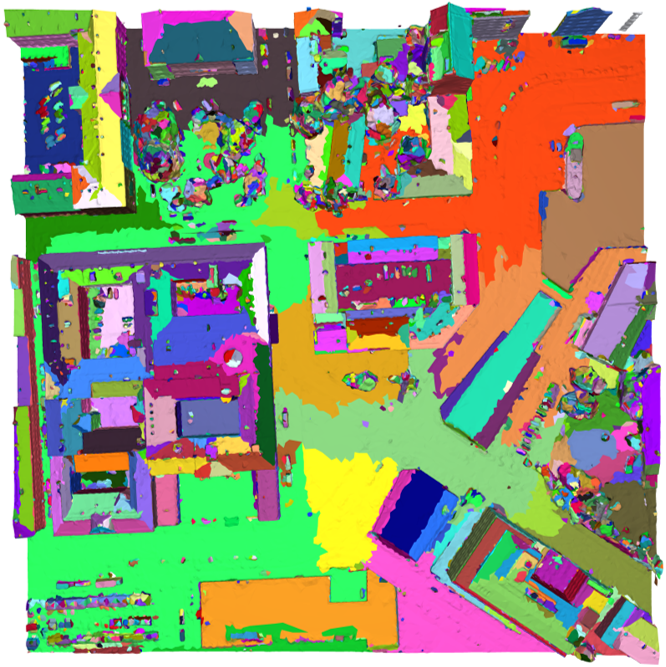}&
			\includegraphics[width=0.25\textwidth]{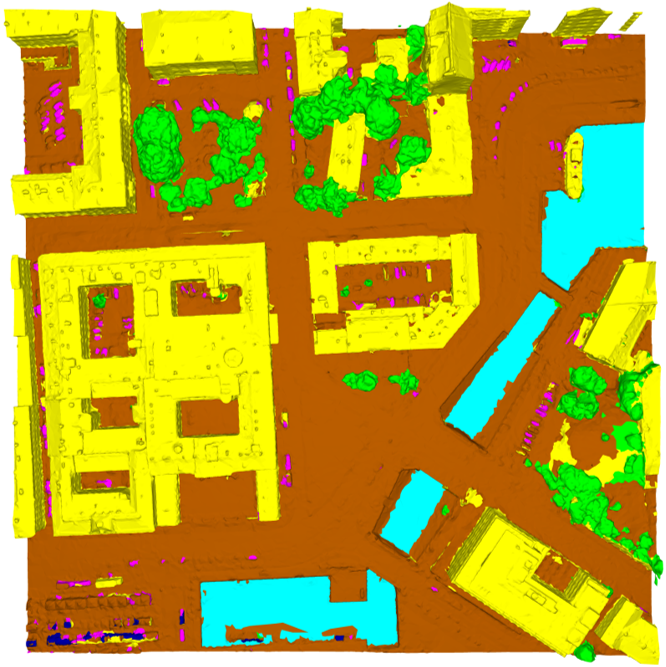}&
			\includegraphics[width=0.25\textwidth]{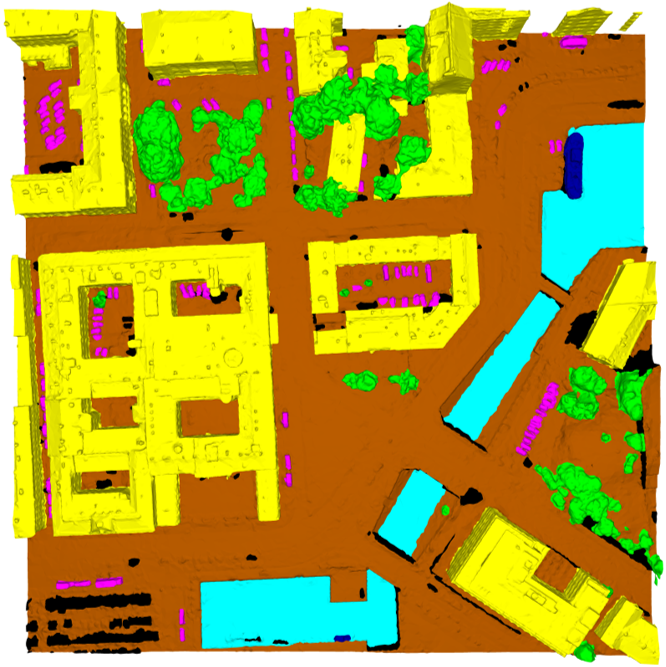}&
			\includegraphics[width=0.25\textwidth]{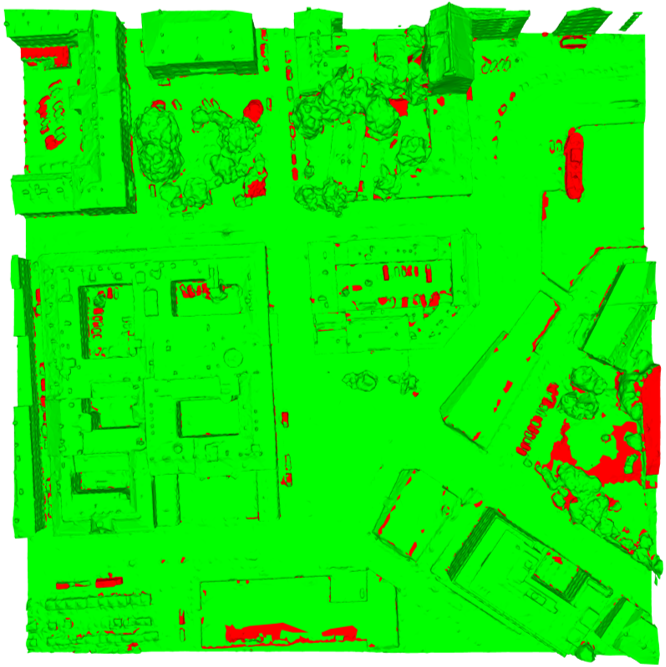}\\		
			\midrule
			\includegraphics[width=0.25\textwidth]{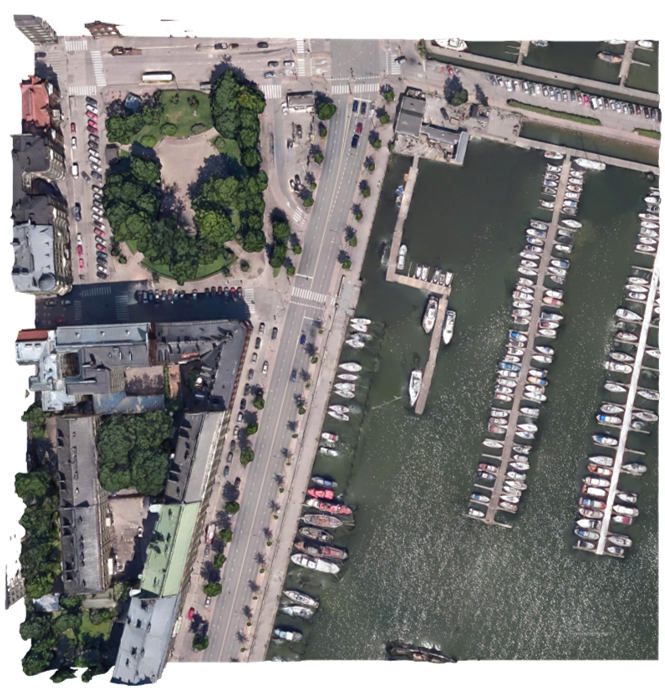}&
			\includegraphics[width=0.25\textwidth]{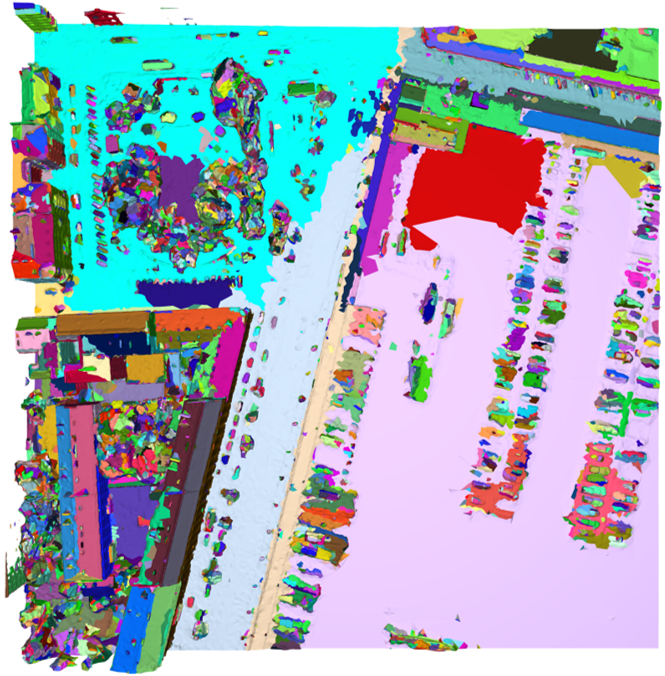}&
			\includegraphics[width=0.25\textwidth]{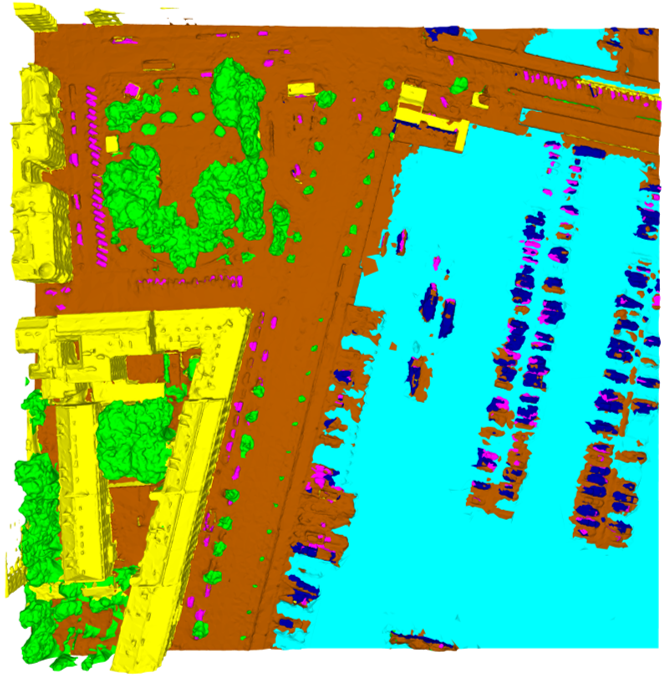}&
			\includegraphics[width=0.25\textwidth]{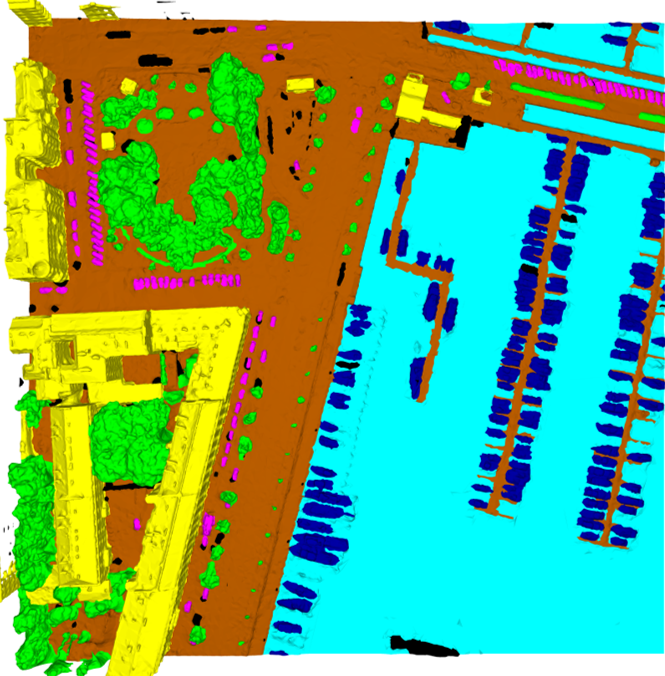}&
			\includegraphics[width=0.25\textwidth]{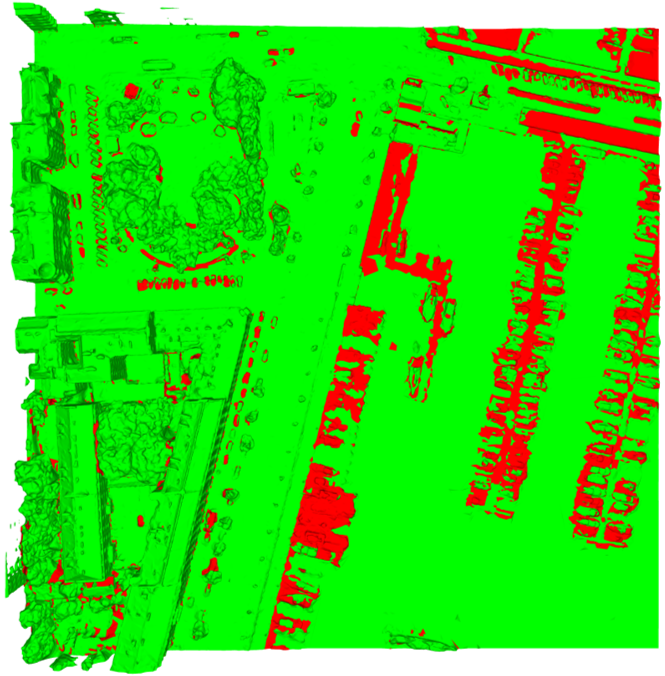}\\	
			\begin{footnotesize} \textbf{(a)} Original \end{footnotesize}&
			\begin{footnotesize} \textbf{(b)} Segments \end{footnotesize}&
			\begin{footnotesize} \textbf{(c)} Predictions \end{footnotesize}&
			\begin{footnotesize} \textbf{(d)} Truth \end{footnotesize}&
			\begin{footnotesize} \textbf{(e)} Error maps \end{footnotesize}
			\\	
		\end{tabular}
	\end{adjustwidth}
	\centering
	\includegraphics[width=0.9\textwidth]{figures/semantic_results/semantic_legend.png}
	\caption{Part of our semantic segmentation results. 
		The first column shows the input texture meshes; 
		the second column shows the over-segmentation results; 
		the third column shows the predicted semantic meshes; 
		the fourth column shows the ground truth meshes; 
		the last column shows the error maps (red: errors; green: correct labels).}
	\label{fig:semantic_qualitive}
\end{figure}

For semantic segmentation, a detailed evaluation of each class is listed in Table \ref{tab:oveal_re}, and we achieve about $ 93.0\% $ overall accuracy and $ 66.2\% $ mIoU as shown in Table \ref{tab:fea_abla}. %
The qualitative evaluation of it is shown in Figure~\ref{fig:semantic_qualitive}.
As shown in Figure~\ref{fig:semantic_qualitive} (e), most of the prediction errors occur at small-scale objects such as vehicles and boats due to fewer training samples and errors from over-segmentation. 

\begin{table}[H]
	\centering
	\noindent\adjustbox{max width=1.0\textwidth}
	{
		\begin{threeparttable}
			\centering
			\begin{tabular}{C{0.4\linewidth}cccc}
				\toprule
				Model & OA (\%) & mAcc (\%) & mIoU (\%) & \textcolor{ao}{$\Delta$mIoU (\%)} \\
				\midrule
				Upper bound (Perfect) & 98.1 & 91.6 & 90.9 & —— \\
				\textbf{Ours (best)} & 93.0 & 70.6 & 66.2 & 0.0 \\
				Without sphericity & 93.0 & 70.5 & 66.1 & -0.1 \\
				Without segment area & 92.9 & 70.5 & 66.0 & -0.2 \\
				Without triangle density & 92.9 & 70.4 & 66.0 & -0.3 \\
				Without variance HSV & 92.9 & 70.3 & 65.9 & -0.3 \\
				Without absolute elevation & 93.0 & 70.2 & 65.9 & -0.3 \\
				Without relative elevation & 92.9 & 70.3 & 65.8 & -0.4 \\
				Without curvature & 92.9 & 70.2 & 65.8 & -0.4 \\
				Without multiscale elevations & 92.8 & 69.8 & 65.1 & -1.1 \\
				Without linearity & 91.8 & 66.6 & 62.0 & -4.2 \\
				Without greenness & 91.9 & 66.6 & 61.9 & -4.3 \\
				Without InMat & 91.6 & 66.4 & 61.6 & -4.6 \\
				Without average HSV & 91.7 & 66.1 & 61.4 & -4.8 \\
				Without verticality & 91.4 & 66.1 & 61.3 & -4.9 \\
				Without HSV histogram bins & 91.5 & 66.0 & 61.1 & -5.1 \\
				\bottomrule
			\end{tabular}%
		\end{threeparttable}
	}
	\caption{Ablation study of the features in our approach.
	The \textit{Upper bound (Perfect)} refers to the maximum achievable performance in theory.}
	\label{tab:fea_abla}
\end{table}%

To better understand the relevance of the features, we measure the feature importance and perform ablation studies (see Table \ref{tab:fea_abla}).
We can observe that the radiometric features (which account for $ 62.8\% $) are more important than geometric ones (which account for $ 37.2\% $).
Moreover, after removing individual feature vectors, the performance will decline, indicating each feature contributes to the best results.

\subsection{Evaluation of Competition Methods}
\begin{figure}[!tb]
	\centering
	\begin{subfigure}[t]{0.32\textwidth}
		\includegraphics[width=\linewidth]{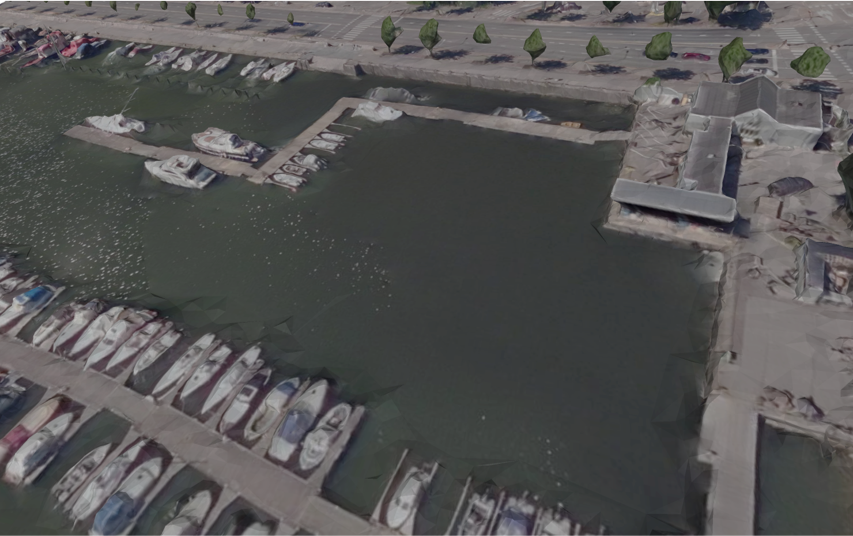}
		\caption{Texture mesh}
	\end{subfigure}
	\hspace*{\fill}
	\begin{subfigure}[t]{0.32\textwidth}
		\includegraphics[width=\linewidth]{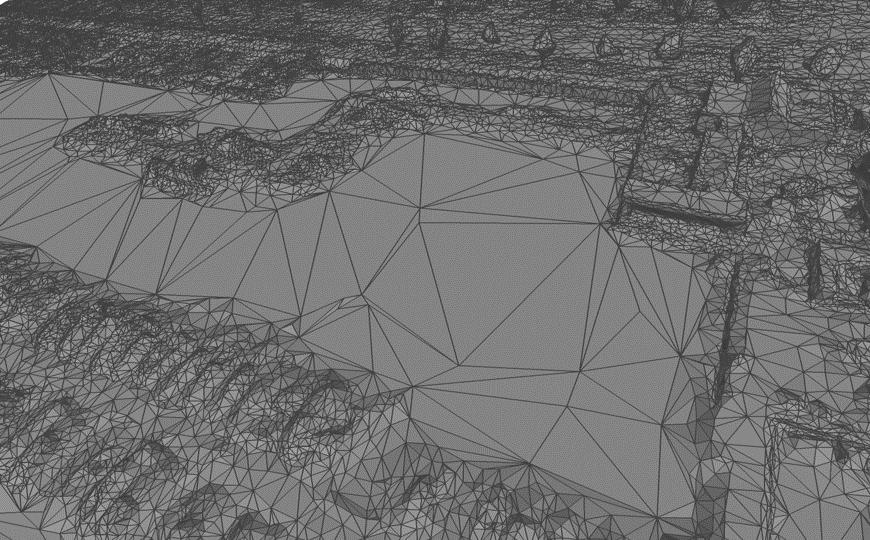}
		\caption{Wireframe}
	\end{subfigure}
	\hspace*{\fill}
	\begin{subfigure}[t]{0.32\textwidth}
		\includegraphics[width=\linewidth]{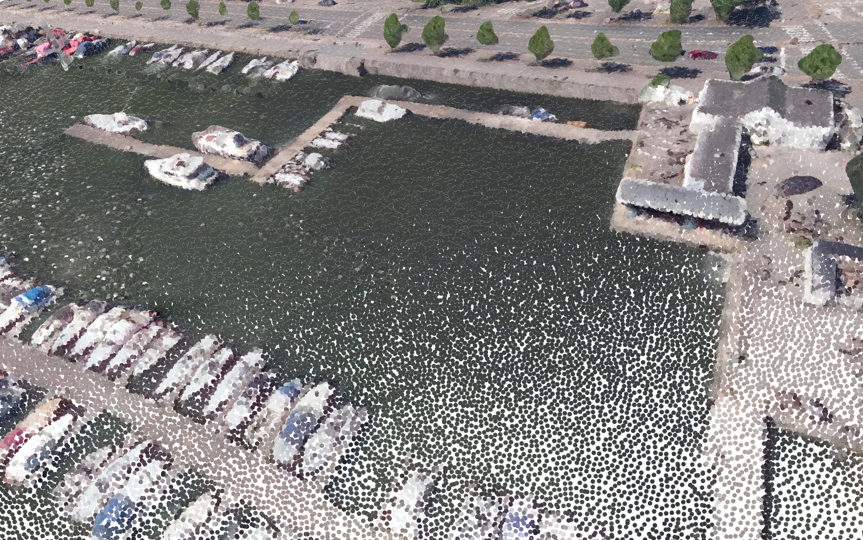}
		\caption{Sampled point cloud}
	\end{subfigure}
	\caption{Sampling point cloud from texture meshes. Our sampled points preserve both geometric and radiometric information of the original mesh.} 
	\label{fig:samplingpts}
\end{figure}

To the best of our knowledge, none of the state-of-the-art deep learning frameworks of 3D semantic segmentation can directly be used on large-scale texture meshes.
Additionally, although the data structures of point clouds and meshes are different, the inherent properties of geometry in the 3D space of the urban environment are nearly identical.
In other words, they can share the feature vectors within the same scenes.
Consequently, we sample the mesh into coloured point clouds (see Figure \ref{fig:samplingpts}) with a density of about 10 $pts/m^2$ as input for the competing deep learning methods.
In particular, we use Montecarlo sampling~\citep{cignoni1998metro} to generate randomly uniform dense samples, and we further prune these samples according to Poisson distributions~\citep{corsini2012efficient} and assign the colour via searching the nearest neighbour from the textures.

To evaluate and compare with the current state-of-the-art 3D deep learning methods that can be applied to a large-scale urban dataset, we select five representative approaches (i.e., PointNet~\citep{qi2017pointnet}, PointNet++~\citep{qi2017pointnet++}, SPG~\citep{landrieu2018large}, KPConv~\citep{thomas2019kpconv}, and RandLA-Net~\citep{hu2020randla}). 
We perform all the experiments on an NVIDIA GEFORCE GTX 1080Ti GPU.
Note that these deep learning-based methods downsample the input point clouds significantly as a pre-processing step. 
In our experiments, the point sampling density is limited by the GPU memory, and increasing or decreasing the sampling density within a reasonable range may lead to slightly different performance.
\textcolor{ao}{
It should be noted that no matter how dense the input point clouds are, almost all state-of-the-art deep learning architectures (such as PointNet, PointNet++, RandLaNet, KPConv, and SPG, etc.) downsample the input point clouds significantly, and they are still able to learn effective features for classification. 
Besides, different deep learning-based point cloud classification frameworks exploit different strategies for downsampling the input points.
}
In addition, we also compare with the joint RF-MRF~\citep{Rouhani2017}, which is the only competition method that directly takes the mesh as input and without using GPU for computation.

The hyper-parameters of all the competing methods are tuned \textcolor{ao}{according to the validation data} to achieve the best results we could acquire.  
Besides, the results of each competitive method (see Table \ref{tab:compare_table}) are demonstrated in average performance based on ten times experiments with the same setting. 
From the comparison results, as shown in Table \ref{tab:compare_table}, we found that our baseline method outperforms other methods except for KPConv.
Specifically, our approach outperforms RF-MRF with a margin of $5.3\%$ mIoU, and deep learning methods (not including KPConv) from $16.7\%$ to $29.3\%$ mIoU.
Compared with the KPConv, the performance of our method is much more robust, which can be observed from Table \ref{tab:compare_table} that the standard deviation of our method is close to zero (i.e., the standard deviation of mIoU of our method is about $0.024\%$).  
The reason is that in our method, we set 100 trees in the random forest to ensure the stability of the model, but in KPConv, the kernel point initialization strategy may not be able to select some parts of the point cloud, which leads to the instability of the results.
Furthermore, compared with all deep learning pipelines, our method is conducted on a CPU and uses much less time for training (including feature computation).
This can be explained by the fact that we have fewer input data (triangles versus points), and the time complexity of our handcrafted features computation is much lower than the features learned from deep learning.

\begin{table}[!tb]
	\noindent\adjustbox{max width=\textwidth}
	{
		\begin{threeparttable}
			\centering
			\begin{tabular}{ccC{0.1\linewidth}ccccccccc}
				\toprule
				 & Terrain & High Vegetation & Building & Water & Vehicle & Boat & mIoU & OA & mAcc & mF1 & \textcolor{ao}{$t_{train}$}\\
				\midrule
				PointNet~\citep{qi2017pointnet} & 56.3 & 14.9   & 66.7 & 83.8 & 0.0 & 0.0 & 36.9 ± 2.3 & 71.4 ± 2.1 & 46.1 ± 2.6 & 44.6 ± 3.2 & 1.8\\
				RandLaNet~\citep{hu2020randla} & 38.9 & 59.6   & 81.5 & 27.7 & 22.0 & 2.1 & 38.6 ± 4.6 & 74.9 ± 3.2 & 53.3 ± 5.1 & 49.9 ± 4.8 & 10.8\\
				SPG~\citep{landrieu2018large} & 56.4 & 61.8 & 87.4   & 36.5 & 34.4 & 6.2 & 47.1 ± 2.4  & 79.0 ± 2.8 & 64.8 ± 1.2 & 59.6 ± 1.9 & 17.8\\
				PointNet++~\citep{qi2017pointnet++} & 68.0 & 73.1   & 84.2 & 69.9 & 0.5 & 1.6 & 49.5 ± 2.1 & 85.5 ± 0.9 & 57.8 ± 1.8 & 57.1 ± 1.7 & 2.8\\
				RF-MRF~\citep{Rouhani2017} & 77.4 & 87.5 & 91.3 & 83.7 & 23.8 & 1.7 & 60.9 ± 0.0 & 91.2 ± 0.0 & 65.9 ± 0.0 & 68.1 ± 0.0 & \textbf{1.1}\\
				KPConv~\citep{thomas2019kpconv} & \textbf{86.5} & 88.4 & \textbf{92.7} & 77.7 & \textbf{54.3} & \textbf{13.3} & \textbf{68.8} ± 5.7 & \textbf{93.3} ± 1.5 & \textbf{73.7} ± 5.4 & \textbf{76.7} ± 5.8 & 23.5\\
				\textbf{Baseline}  & 83.3          & \textbf{90.5} & 92.5 & \textbf{86.0} & 37.3 & 7.4 & 66.2 ± 0.0 & 93.0 ± 0.0 & 70.6 ± 0.0 & 73.8 ± 0.0 & 1.2\\
				\bottomrule
			\end{tabular}%
		\end{threeparttable}
	}
	\caption{Comparison of various semantic segmentation methods on the new benchmark dataset. 
		The results reported in this table are per-class IoU (\%), mean IoU (mIoU, \%) ± standard deviation, Overall Accuracy (OA, \%) ± standard deviation, mean class Accuracy (mAcc, \%) ± standard deviation, mean F1 score (mF1, \%) ± standard deviation, and the time cost of training (\textcolor{ao}{$t_{train}$}, hours). 
		The running times of SPG include both feature computation and graph construction, and RF-MRF and our baseline method include feature computation.
		We repeated the same experiment ten times and presented the mean performance.} 
	\label{tab:compare_table}
\end{table}%

\subsection{Evaluation of Annotation Refinement}
Following the proposed framework, a total of 19,080,325 triangle faces have been labelled, which took around 400 working hours.
Compared with a triangle-based manual approach, we estimate that our framework saved us more than 600 hours of manual labour.
Specifically, we have measured the labelling speed with these two different approaches on the same mesh tile consisting of 309,445 triangle faces and 8,033 segments.
It took around 17 hours for manual labelling based on triangle faces, while with our segment-based semi-automatic approach, it took only 6.5 hours.  

\begin{figure}[!tb]
	\centering
	\begin{subfigure}[t]{0.31\textwidth}
		\includegraphics[width=\linewidth]{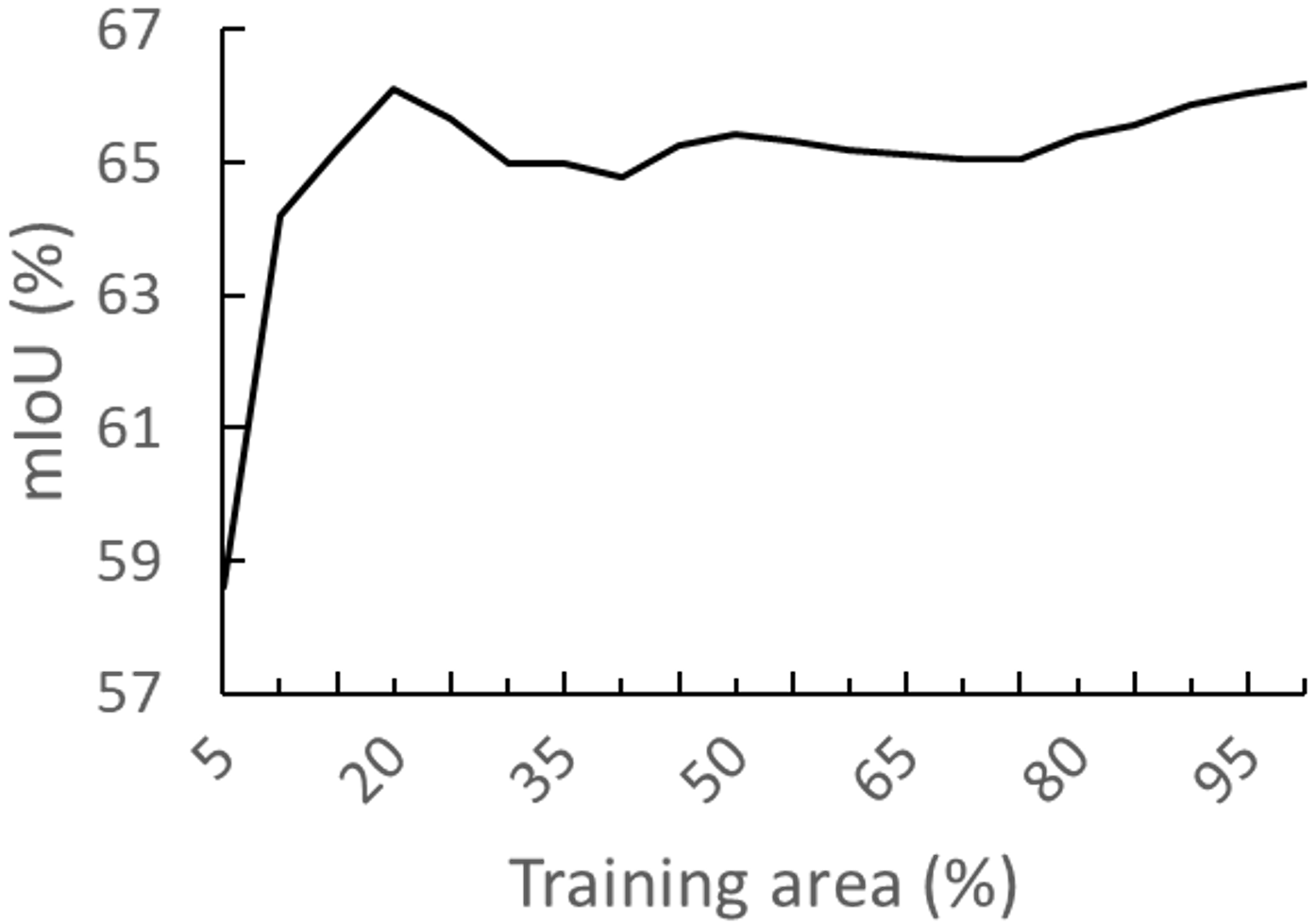}
		\caption{\textcolor{ao}{mIoU}}
		\label{fig:tr_area_miou}
	\end{subfigure}
	\hspace*{\fill}
	\begin{subfigure}[t]{0.31\textwidth}
		\includegraphics[width=\linewidth]{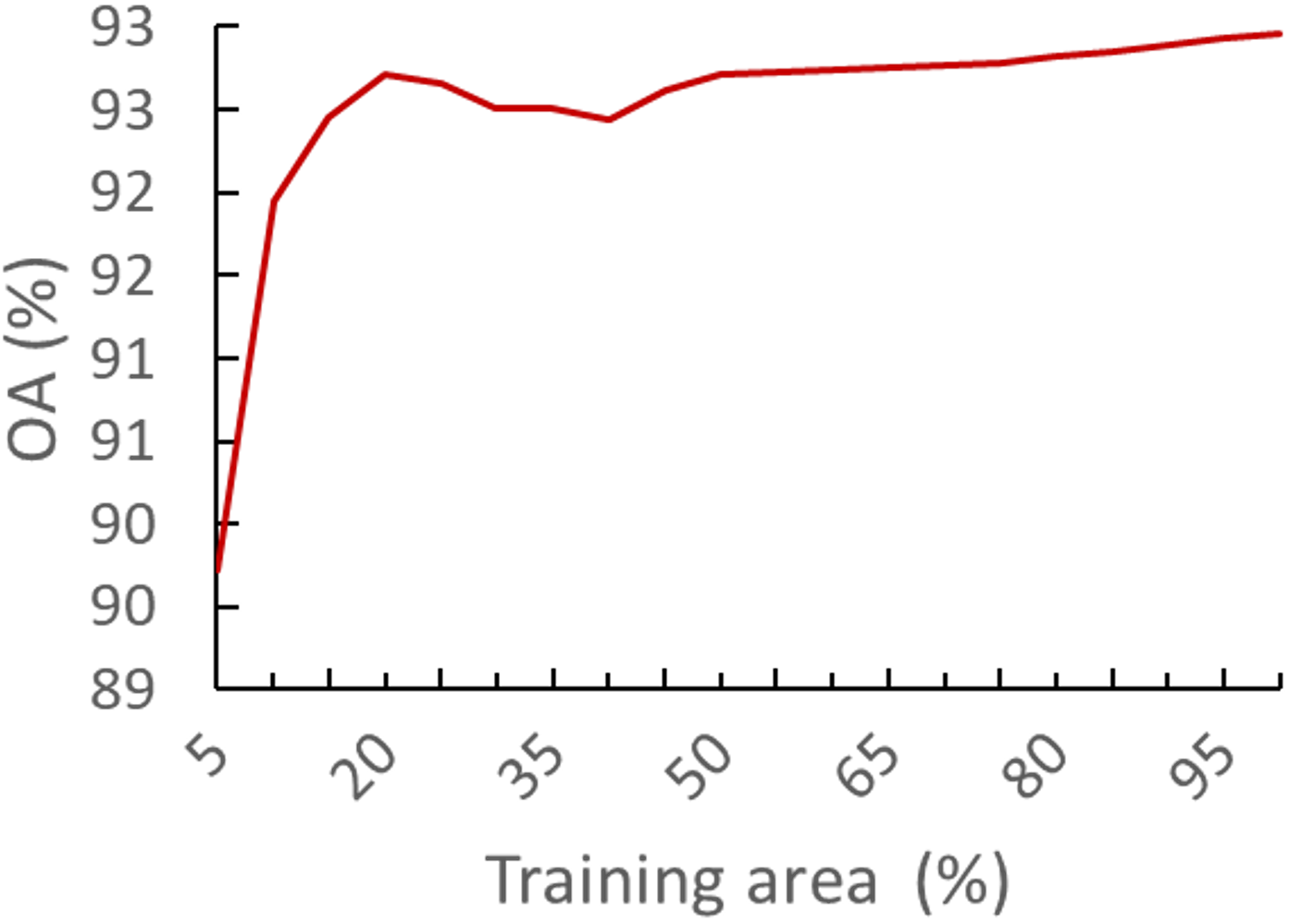}
		\caption{\textcolor{ao}{OA}}
		\label{fig:tr_area_oa}
	\end{subfigure}
	\hspace*{\fill}
	\begin{subfigure}[t]{0.32\textwidth}
		\includegraphics[width=\linewidth]{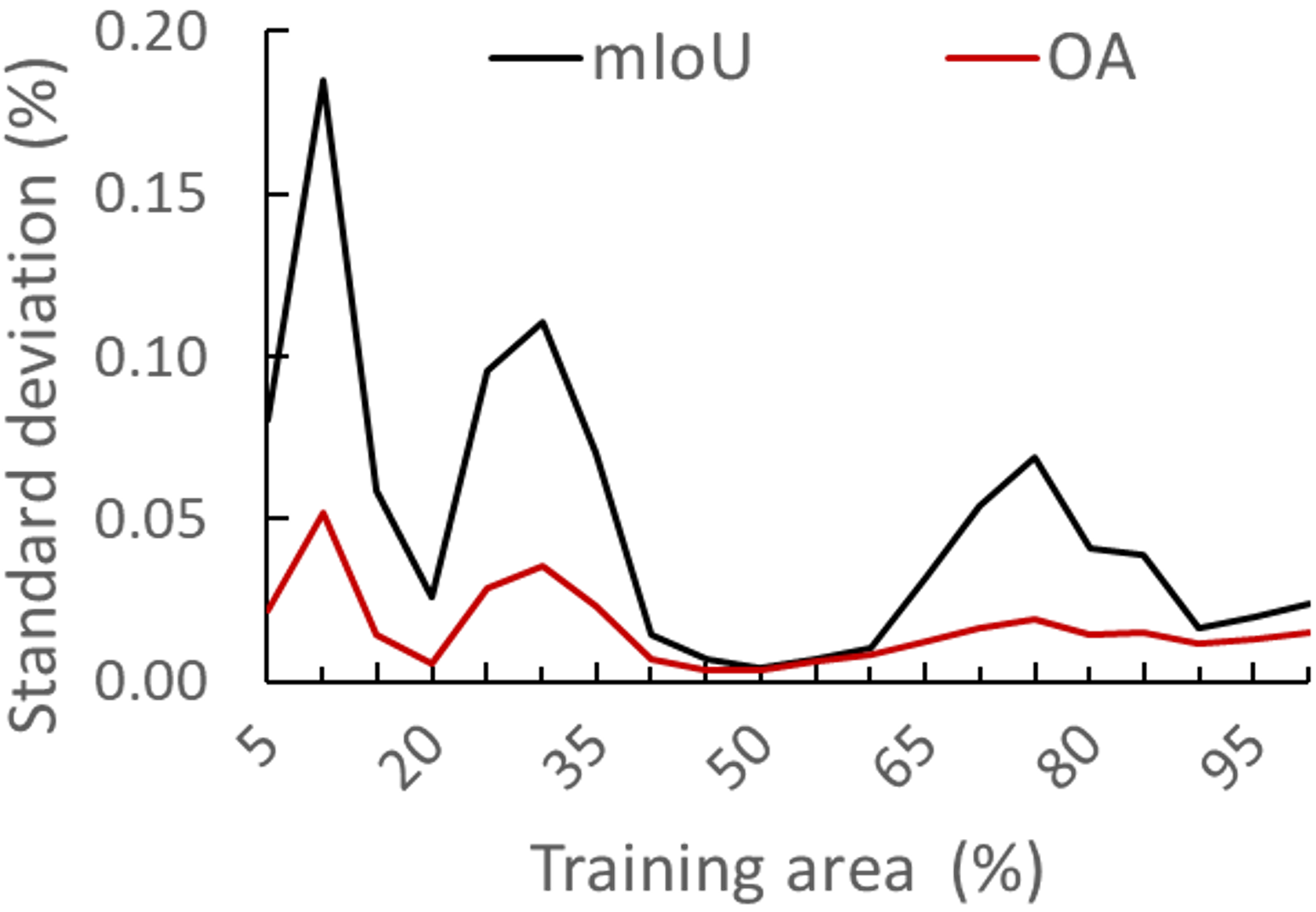}
		\caption{\textcolor{ao}{Standard deviation}}
		\label{fig:tr_area_std}
	\end{subfigure}
	\caption{Effect of the amount of training data on the performance of the initial segmentation method used in the semi-automatic annotation. We repeated the same experiment ten times for each set of training areas and presented the mean performance.}
	\label{fig:tr_area}
\end{figure}

We also evaluate the performance of semantic segmentation with different amounts of input training data on our baseline approach with the intention of understanding the required amount of data to obtain decent results. 
Specifically, we use ten sets of different training areas with ten times experiments with the same configuration of each set, and we linearly interpolate the results as shown in Figure \ref{fig:tr_area}. 
From Figures \ref{fig:tr_area_miou}, \ref{fig:tr_area_oa}, and \ref{fig:tr_area_std}, we can observe that our initial segmentation method only requires about $10\%$ (equal to about 0.325 $km^2$) of the total training area to achieve acceptable and stable results. 
In other words, using a small amount of ground truth data, our framework can provide robust pre-labelled results and significantly reduce the manually labelling efforts.

	\section{Conclusion}\label{sec:Discussion}
We have developed a semi-automatic mesh annotation framework to generate a large-scale semantic urban mesh benchmark dataset covering about 4 $km^2$. 
In particular, we have first used a set of handcrafted features and a random forest classifier to generate the pre-labelled dataset, which saved us around 600 hours of manual labour. 
Then we have developed a mesh labelling tool that allows the users to interactively refining the labels at both the triangle face and the segment levels.
We have further evaluated the current state-of-the-art semantic segmentation methods that can be applied to large-scale urban meshes, and as a result, we have found that our classification based on handcrafted features achieves $93.0\%$ overall accuracy and $66.2\%$ of mIoU.
This outperforms the state-of-the-art machine learning and most deep learning-based methods that use point clouds as input.
Despite this, there is still room for improvement, especially on the issues of imbalanced classes and object scalability. 
For future work, we plan to label more urban meshes of different cities and extend our Helsinki dataset to include parts of urban objects (such as roof, chimney, dormer, and facade). 
We will also investigate smart annotation operators (such as automatic boundary refinement and structure extraction), which involve more user interactivity and may help reduce further the manual labelling task.
	
	\section{Acknowledgements}\label{sec:acknowledgements} 
We would like to thank EuroSDR for providing the funding for this project.
The authors appreciate the people who have helped the project, especially Ziqian Ni for the development and the testing of the annotation platform, and Mels Smit and Charalampos Chatzidiakos for assisting with the annotation of the meshes.

	\bibliography{mybibfile}
	
\end{document}